\documentclass[11pt]{article}

\PassOptionsToPackage{table}{xcolor}

\usepackage[final]{acl}

\usepackage{times}
\usepackage{latexsym}
\usepackage{booktabs}
\usepackage{multirow}
\usepackage[T1]{fontenc}

\usepackage[utf8]{inputenc}

\usepackage{microtype}

\usepackage{amsmath}

\usepackage{inconsolata}

\usepackage{graphicx}

\usepackage{contour}

\usepackage[most]{tcolorbox}
\usepackage{fancyvrb}
\usepackage{fvextra}
\usepackage{caption}
\usepackage{float}
\tcbuselibrary{breakable,skins}
\usepackage{capt-of}
\usepackage{xcolor}
\usepackage{epstopdf}

\usepackage{arabtex}
\usepackage{utf8}

\newcommand{\ArabiGEE}{\textbf{ArabiGEE}}
\newcommand{\hide}[1]{}

\newcommand{\AHAMZAUP}{{\^{A}}}
\newcommand{\WHAMZA}{{\^{w}}}
\newcommand{\AHAMZADN}{{\v{A}}}
\newcommand{\YHAMZA}{{\^{y}}}

\title{ArabiGEE: A Hierarchical Taxonomy \\ for Arabic Grammatical Error Explanation}

\author{Khaled Elhady\textsuperscript{1*} \quad Omar Kallas\textsuperscript{1*} \quad \textbf{Nizar Habash\textsuperscript{1,2}} \quad Bashar Alhafni\textsuperscript{1}\\ \textsuperscript{1}Mohamed bin Zayed University of Artificial Intelligence \\ \textsuperscript{2}New York University Abu Dhabi\\ \texttt{\{khaled.marzouk,omar.kallas,bashar.alhafni\}@mbzuai.ac.ae}\\ \texttt{nizar.habash@nyu.edu}}

\begin{document}
\setcode{utf8}
\vocalize


\maketitle

\begingroup
\renewcommand{\thefootnote}{*}
\NoHyper
\footnotetext{Equal contribution.}
\endNoHyper
\endgroup

\begin{abstract}
We introduce {\ArabiGEE}, the first hierarchical Arabic grammatical error explanation (GEE) taxonomy grounded in explicit error types. Unlike existing GEE approaches that treat explanation generation as free-form text, {\ArabiGEE} organizes grammatical explanations hierarchically across orthographic, morphological, syntactic, and lexical dimensions while capturing interactions across them. The taxonomy comprises 28 error types, 177 correction types, and 374 explanation types, supporting 2,218 explanation groups. We apply {\ArabiGEE} to manually annotate 1,000 erroneous-corrected word pairs sampled from Arabic grammatical error correction (GEC) corpora, producing 1,232 groups comprising 1,439 explanation instances. We show how our taxonomy enables framing GEE as a structured prediction task, allowing automatic evaluation of LLMs on Arabic GEE. Our code and data are publicly available.\footnote{\url{https://github.com/mbzuai-nlp/arabigee}}


\end{list} 
\end{abstract}
 
\section{Introduction}
Grammatical error correction (GEC) has important applications in writing assistance and language learning. While GEC has witnessed significant progress, particularly for English \cite{bryant-etal-2023-grammatical}, most systems generate corrections without explaining the underlying grammatical phenomena. Such explanations are particularly valuable for language learning and pedagogy because they help learners understand why a correction is needed \cite{wakabayashi2008effect,kang2015efficacy,brown2023Effectivenessow,williams2024delivering,lu2025feedback}.

With the rise of large language models (LLMs), recent work has explored grammatical error explanation (GEE) \cite{song-etal-2024-gee}, where models generate natural language explanations for errors given an erroneous sentence and its correction. Existing work largely treats GEE as a free-form generation task without grounding explanations in explicit error types \cite{song-etal-2024-gee,kaneko-okazaki-2024-controlled,lopez-cortez-etal-2024-gmeg}, making automatic evaluation difficult since generation metrics cannot reliably determine whether a model correctly explains errors.

\begin{figure}[t!]
    \centering
    \includegraphics[width=\linewidth]{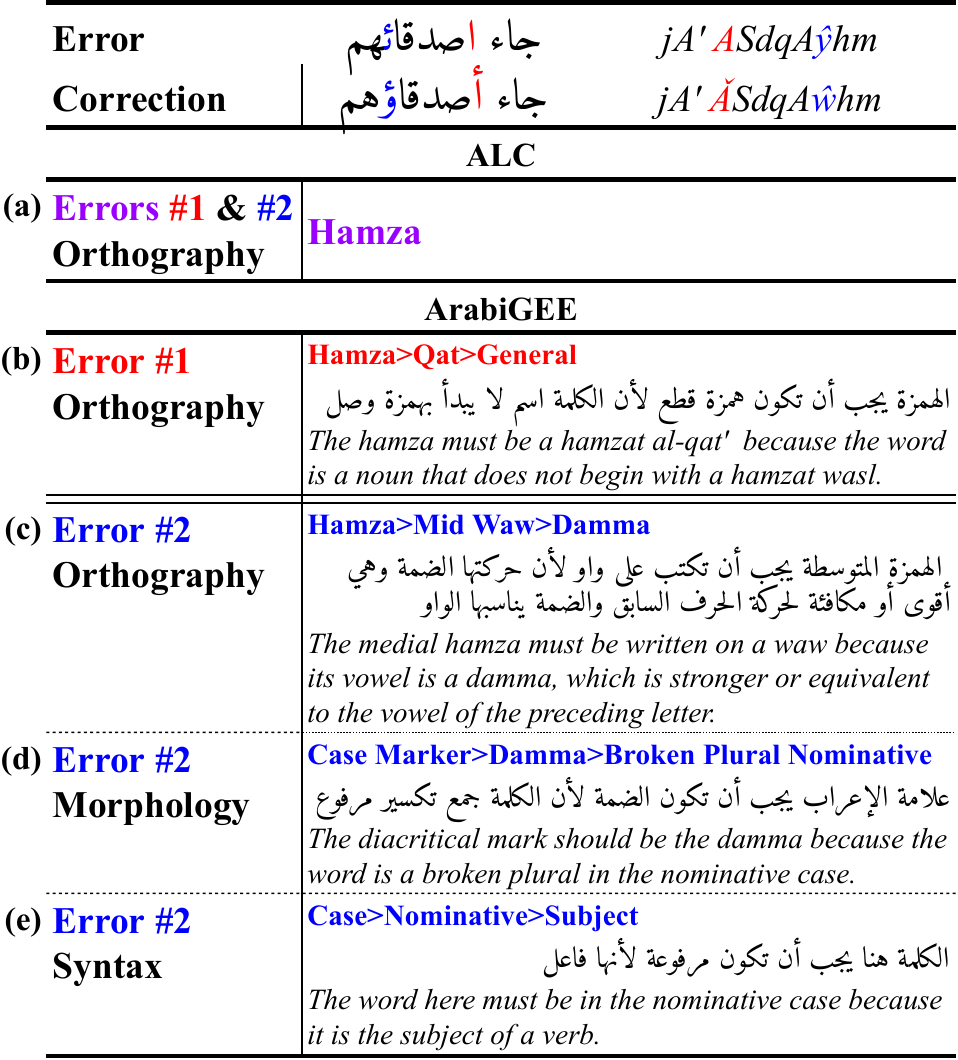}
    \caption{Example comparing flat error typing in the ALC taxonomy with hierarchical explanations in {\ArabiGEE}. The sentence translates to \textit{``their friends came''}.}
    \label{fig:intro-example}
\end{figure}

GEE research has focused largely on English, while morphologically rich languages such as Arabic have received little attention. Recent Arabic GEE efforts \cite{magdy-etal-2024-gazelle,mubarak-etal-2026-nahw} follow the same free-form formulation without grounding explanations in explicit error types. Existing Arabic error taxonomies \cite{Alfaifi:2013:error,alfaifi-atwell-2014-evaluation} further complicate this problem, as they rely on flat error typing that cannot effectively ground grammatical explanations or distinguish multiple instances of the same error type within a word (Figure \ref{fig:intro-example}(a)). Moreover, many Arabic errors involve interacting orthographic, morphological, syntactic, and lexical phenomena, requiring multiple complementary explanations across linguistic dimensions (Figure \ref{fig:intro-example}(c--e)) to fully explain why a correction is needed.

To address these limitations, we introduce {\ArabiGEE}, a manually designed \textbf{Arabi}c \textbf{GEE} taxonomy grounded in explicit error types. {\ArabiGEE} organizes grammatical explanations hierarchically into 28 error types, 177 correction types, and 374 explanation types, supporting 2,218 explanation groups. Rather than treating explanations as free-form text, our framework links them to structured linguistic phenomena and their correction patterns, allowing us to formulate GEE as a structured prediction task. We apply {\ArabiGEE} to manually annotate 1,000 erroneous-corrected word pairs sampled from three Arabic GEC corpora covering native speaker (L1) and second-language learner (L2) errors, producing 1,232 groups comprising 1,439 explanation instances. Moreover, we evaluate diverse LLMs, including multilingual models and an Arabic-centric model, and analyze their performance across linguistic dimensions and input variants. Our contributions are summarized as follows:

\begin{enumerate}
    \item We introduce the first hierarchical, manually designed Arabic GEE taxonomy.
    \item We manually annotate 1,000 word pairs from three Arabic GEC corpora using {\ArabiGEE}, producing 1,232 groups comprising 1,439 explanation instances.
    \item We show how {\ArabiGEE} can be used to automatically evaluate LLMs on Arabic GEE.
\end{enumerate}

\section{Background and Related Work}
\subsection{Error Type Annotation}
\label{sec:related-work-error-typing}
Research on GEC has led to the development of several error taxonomies \cite{yannakoudakis-etal-2011-new,ng-etal-2013-conll,dahlmeier-etal-2013-building,ng-etal-2014-conll,boyd-etal-2014-merlin,rozovskaya-roth-2019-grammar,zhang-etal-2022-mucgec,naplava-etal-2022-czech}, enabling the development of automatic error type annotation tools widely used in GEC evaluation \cite{bryant-etal-2017-automatic,choshen-etal-2020-classifying}.  Closest to our work, \newcite{coyne-2025-annotating} introduced a hierarchical error typology for L2 errors in English.



In Arabic, the primary effort on error type annotation is the Arabic Learner Corpus (ALC) taxonomy introduced by \newcite{alfaifi-atwell-2012-arabic,alfaifi-atwell-2014-evaluation}. The taxonomy consists of orthographic, morphological, syntactic, and semantic dimensions, and later served as the foundation for ARETA \cite{belkebir-habash-2021-automatic}, an automatic error type annotation tool for Arabic. However, the taxonomy was primarily designed for flat error typing rather than structured grammatical explanation. While it can assign multiple error types within the same word, it cannot distinguish multiple instances of the same error type, and each error is assigned to a single linguistic dimension. In contrast, \textbf{our work} grounds grammatical explanations in explicit error and correction types through a hierarchical cross-dimensional GEE framework (\S\ref{sec:arabigee}).



\subsection{Grammatical Error Explanation}
Beyond error typing, there has been growing interest in generating explanations to help learners understand grammatical revisions \cite{nagata-2019-toward,nagata-etal-2020-creating,nagata-etal-2021-shared}, particularly with the rise of LLMs. Recently, \newcite{song-etal-2024-gee} formalized GEE and evaluated LLMs on it, with further studies exploring LLMs for GEE \cite{kaneko-okazaki-2024-controlled,lopez-cortez-etal-2024-gmeg,maity-2024-how,vainikko-etal-2025-paragraph,ye-2025-excgec,coyne-2025-annotating}. However, prior work treats GEE as a free-form text generation task, evaluated using surface-overlap metrics that cannot verify explanation correctness, or with human and LLM-based judgment, which is costly and hard to reproduce at scale.





Only two prior efforts have explored Arabic GEE. \newcite{magdy-etal-2024-gazelle} introduced Gazelle, an Arabic instruction dataset that extends the ALC taxonomy with finer-grained error subclasses and associated explanations sourced from online educational resources, though without explicitly modeling them within a structured taxonomy or verifying explanation correctness. Similarly, \newcite{mubarak-etal-2026-nahw} introduced Nahw, a dataset for evaluating LLMs on GEE, where explanations were collected from online resources and manually verified by linguists, but without grounding them in explicit error types. Both efforts treat GEE as a free-form generation task, and neither applies explanations to existing Arabic GEC corpora, limiting systematic evaluation across naturally occurring errors. 

\textbf{In our work}, we introduce a manually designed hierarchical Arabic GEE taxonomy and apply it to existing L1 and L2 Arabic GEC datasets through manual annotation. We show how our taxonomy enables framing GEE as a structured prediction task, supporting automatic evaluation. To our knowledge, we are the first to frame GEE this way.




\subsection{Arabic Linguistic Challenges}
\label{sec:arabic-facts}
Arabic exhibits a diglossic linguistic situation \cite{Ferguson:1959:diglossia}, where Modern Standard Arabic (MSA), the standardized variety used in education and formal settings, coexists with unstandardized regional dialects that are primarily spoken. As no Arabic speaker acquires MSA natively, writing in MSA is effectively a second-language task even for native speakers, often resulting in dialectal code-switching across orthographic, morphological, syntactic, and lexical levels \cite{Habash:2008:guidelines}. Our work focuses on MSA.

While MSA orthography is standardized, written MSA still exhibits substantial orthographic inconsistencies even in professionally written text \cite{habash-etal-2012-conventional}. In addition, Arabic has a rich morphological system that inflects for numerous features \cite{Habash:2010:introduction}. These properties often lead to errors whose correction depends on multiple interacting linguistic phenomena simultaneously. For example, a single correction may involve orthographic changes whose explanation depends on morphological or syntactic information.

Figure \ref{fig:intro-example} illustrates these challenges. The erroneous word \<اصدقائهم> \textit{ASdqA{\YHAMZA}hm}\footnote{Arabic HSB transliteration \cite{Habash:2007:arabic-transliteration}.} contains two distinct Hamza-related (glottal stop) errors. However, flat or single-layer error taxonomies such as the ALC assign both under a single Hamza label within the orthographic dimension (Figure \ref{fig:intro-example}(a)), making it difficult to represent distinct correction patterns and interacting linguistic phenomena. In contrast, our {\ArabiGEE} taxonomy identifies the two errors separately (Figure \ref{fig:intro-example}(b) and (c--e)) and models how a single correction may require explanations across multiple interacting dimensions, including orthography, morphology, and syntax (Figure \ref{fig:intro-example}(c--e)).
The second error is particularly illustrative of these interactions. In Arabic, the orthographic form of the Hamza character \{\<ءأإؤئ>\} \{\textit{' {\AHAMZAUP \AHAMZADN \WHAMZA \YHAMZA}}\} is sensitive to its surrounding vocalic environment. In the erroneous form, the Hamza is written on \<ئ>~\textit{\YHAMZA} as if it were preceded by or associated with the vowel \textit{/i/}, which also corresponds to the genitive case marker. In the corrected form, however, the Hamza appears on \<ؤ>~\textit{\WHAMZA}, reflecting the vowel \textit{/u/} associated with the nominative marker. Crucially, this nominative assignment is itself determined syntactically by the governing verb in the example context. Thus, fully explaining the correction requires all three layers simultaneously: orthography to explain the Hamza form, morphology to explain the underlying case/vowel realization, and syntax to explain why nominative case is assigned in the first place. 
Some of this information is contextual and not directly visible in undiacritized Arabic writing, further motivating cross-dimensional explanations.


\section{Arabic GEE Taxonomy}
\label{sec:arabigee}

\subsection{Taxonomy Design}
\label{sec:arabigee-design}
{\ArabiGEE} aims to capture the linguistic complexity of Arabic grammatical errors (\S\ref{sec:arabic-facts}) by organizing explanations hierarchically in linguistic dimensions and modeling dependencies among them.

\paragraph{Linguistic Dimensions}
Our taxonomy is organized into four linguistic dimensions, each capturing a different class of grammatical errors together with the linguistic information needed to explain them. The \textit{Lexical} (Lex) dimension captures errors related to word choice and clitic usage, including dialectal code-switching into MSA. The \textit{Orthographic} (Orth) dimension models surface-related phenomena such as letter confusion, insertion, or deletion. The \textit{Morphological} (Morph) dimension explains changes involving morphological features such as case, gender, number, and state markers. Finally, the \textit{Syntactic} (Synt) dimension captures errors whose explanation depends on the context and syntactic relations between words, such as assignment, agreement, and syntactic function.

\paragraph{Hierarchical Structure} Each dimension is structured hierarchically into three levels: \textit{Error Types}, \textit{Correction Types}, and \textit{Explanation Types}. The dimensions and error types form the underlying error taxonomy, while the full hierarchy defines the GEE taxonomy. \textit{Error Types} describe classes of errors, such as spelling, case, or gender errors. \textit{Correction Types} capture correction patterns associated with an error type, including substitutions, insertions, and deletions, e.g., modifying letter forms or assigning nominative case. \textit{Explanation Types} provide fine-grained linguistic specifications of the correction. Error, correction, and explanation types are not restricted to one-to-one word edits: they capture one-to-many (e.g., space insertion) and many-to-one (e.g., space deletion) corrections.


\paragraph{Explanation Paths and Groups}
Within each linguistic dimension, an error is explained by traversing the hierarchy from its error type through its correction type to its explanation type. We refer to this sequence as an \textit{Explanation Path}, which maps to a single fixed, learner-facing \textit{Explanation Text}. We define an \textit{Explanation Group} as the set of explanation paths jointly explaining an individual error, with at most one path per linguistic dimension. An explanation group is thus either \textit{single-dimensional}, consisting of a single path, or \textit{cross-dimensional}, consisting of multiple paths from distinct dimensions. A word pair may contain multiple explanation groups, each corresponding to a distinct error.


\paragraph{Cross-Dimensional Dependencies} Many Arabic grammatical errors cannot be fully explained within a single linguistic dimension (\S\ref{sec:arabic-facts}). For instance, some orthographic explanations depend on morphological information, such as when the correct form of a Hamza is determined by the underlying vowel. Similarly, some morphological explanations depend on syntactic structure, such as when nominative case marking is determined by whether a word functions as a subject. To capture these dependencies, our taxonomy supports explanation groups spanning three combinations of linguistic dimensions: orthographic--morphological, morphological--syntactic, and orthographic--morphological--syntactic. Explanation paths within a group must be linguistically compatible. Lexical explanation paths do not participate in cross-dimensional groups.

\begin{figure*}[t!]
    \centering
    \includegraphics[width=\textwidth]{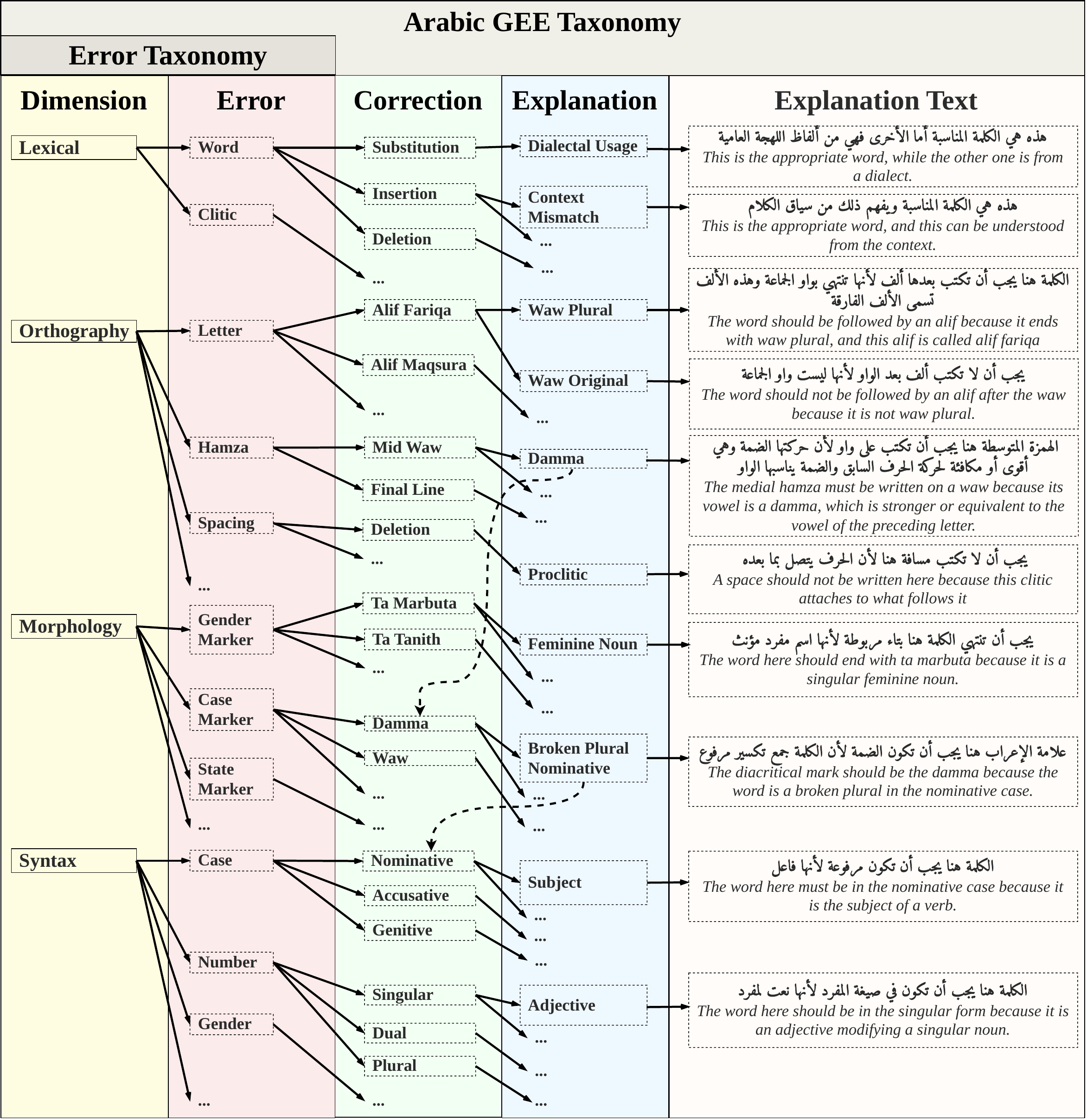}
    \caption{Overview of the {\ArabiGEE} taxonomy, showing hierarchical explanation paths within dimensions (solid arrows) and cross-dimensional links between related linguistic layers (dashed arrows).}
    \label{fig:taxonomy-design}
\end{figure*}

Figure~\ref{fig:taxonomy-design} illustrates the overall structure of {\ArabiGEE}.

\begin{table}[t]
\centering
\begin{tabular}{lccc}
\toprule
\textbf{Dimension(s)} & \textbf{Err.} & \textbf{Corr.} & \textbf{Exp.} \\
\midrule
\rowcolor{gray!15}
\multicolumn{4}{c}{\textit{Single-dimensional}} \\
Lex & 2 & 6 & 10 \\
Orth & 5 & 46 & 85 \\
Morph & 14 & 107 & 157 \\
Synt & 7 & 18 & 122 \\
\textbf{Subtotal} & \textbf{28} & \textbf{177} & \textbf{374} \\
\midrule
\rowcolor{gray!15}
\multicolumn{4}{c}{\textit{Cross-dimensional}} \\
Orth+Morph & 5 & 51 & 122 \\
Morph+Synt & 8 & 94 & 1,057 \\
Orth+Morph+Synt & 2 & 33 & 665 \\
\textbf{Subtotal} & \textbf{15} & \textbf{178} & \textbf{1,844} \\
\midrule
\textbf{Total} & \textbf{43} & \textbf{355} & \textbf{2,218} \\
\bottomrule
\end{tabular}
\caption{Number of distinct combinations of error (Err.), correction (Corr.), and explanation (Exp.) types supported by {\ArabiGEE} for single- and cross-dimensional explanation groups.}
\label{tab:taxonomy-stats}
\end{table}


\subsection{Taxonomy Construction}
The taxonomy was manually developed by two co-authors of this paper, both native Arabic speakers with formal training in Arabic grammar, linguistics, and educational NLP. Its construction involved iterative analysis of a standard Arabic grammar reference \cite{dahdah:1987:arabic-grammar}, Arabic GEC examples, and the ALC taxonomy to identify recurring learner errors and correction patterns. While some high-level error types overlap with the ALC taxonomy, {\ArabiGEE} was independently designed to support hierarchical grammatical explanations and cross-dimensional linguistic modeling (\S\ref{sec:related-work-error-typing}). We then manually defined explanation paths and corresponding learner-facing texts that are linguistically precise and pedagogically meaningful.

Table~\ref{tab:taxonomy-stats} summarizes our taxonomy's coverage. Single-dimensional counts reflect individual types, with the explanation-level subtotal counting explanation paths (single-dimensional groups). Cross-dimensional counts reflect valid explanation path combinations and their distinct projections onto correction and error type levels. The explanation-level total counts all supported groups.


\begin{table}[t]
\centering
\setlength{\tabcolsep}{6pt}
\begin{tabular}{lcccc}
\toprule
\shortstack{\textbf{ALC Error}} & \textbf{Q14} & \textbf{Q15} & \textbf{ZAE} & \textbf{Total} \\\midrule
{Orthography} & 342 & 147 & 33 & 522 \\
{Syntax}      & 189 & 96  & 24 & 309 \\
{Semantic}    & 58  & 35  & 6  & 99 \\
{Morphology}  & 41  & 37  & 6  & 84  \\
{Unknown}     & 24  & 12  & 3  & 39  \\
\bottomrule
\end{tabular}
\caption{Distribution of ARETA-assigned ALC errors in the sampled word pairs.}
\label{tab:sampled-areta-category-distribution}
\end{table}

\subsection{Data Selection}
\label{sec:data-selection}

To demonstrate {\ArabiGEE}'s usefulness, we apply it to annotate portions of three publicly available Arabic GEC development sets covering L1/L2 learner errors: QALB-2014 (Q14; L1) \cite{mohit-etal-2014-first}, QALB-2015 (Q15; L2) \cite{rozovskaya-etal-2015-second}, and ZAEBUC (ZAE; L1) \cite{habash-palfreyman-2022-zaebuc}. Alongside parallel erroneous and corrected sentences, the datasets contain word-level alignment and error type annotations produced by ARETA \cite{belkebir-habash-2021-automatic} following the ALC taxonomy. These error types serve only as proxies for diverse error sampling, without constraining error counts or types identified using {\ArabiGEE}. Dataset statistics are in Appendix~\ref{app:app-gec-data-stats}.

We sample 1,000 erroneous-corrected word pairs from the three datasets. Sampling was designed for broad coverage across ALC error categories: orthography, syntax, semantic, morphology, and uncategorized errors (unknown).\footnote{ALC's semantic category corresponds to our Lexical dimension, extended to other clitics and insertions/deletions.} While most of the sampled word pairs are associated with a single ALC error type, a subset involves multiple distinct error types within the same word pair. Table~\ref{tab:sampled-areta-category-distribution} summarizes the distribution of sampled word pairs across ALC error categories; counts slightly exceed 1,000 since some pairs carry multiple ALC errors.

\subsection{Annotation}

The same two annotators who developed the {\ArabiGEE} taxonomy independently annotated the 1,000 sampled word pairs using the proposed taxonomy. The annotators were not financially compensated for their work. Given each aligned word pair in both erroneous and corrected contexts, the annotators identified the individual errors and assigned to each error an explanation group consisting of one or more explanation paths (i.e., Error Type $\rightarrow$ Correction Type $\rightarrow$ Explanation Type), along with their corresponding learner-facing explanation texts.

\subsection{Inter-annotator Agreement}
We assess agreement at the word-pair level by aligning the explanation groups assigned by the two annotators using bipartite matching, which maximizes agreement across taxonomy levels independently of the order of the groups (Appendix~\ref{app:matching}). Two annotations agree exactly if they contain identical explanation groups, irrespective of their order. Under this strict criterion, the annotators achieved 78.7\% exact agreement. All disagreements were subsequently resolved jointly to produce the final gold annotations.

Among the 213 disagreeing word pairs, 23.5\% exhibit disagreements in the number of identified errors. Across all four linguistic dimensions, particularly orthography, annotators disagree more frequently on the detection of dimensions than on the assignment of types within shared dimensions. We provide a detailed breakdown of disagreements in Appendix~\ref{app:iaa-disagreements}.

Across all 1,000 word pairs, we further measure agreement at each taxonomy level within each dimension using accuracy and macro-averaged precision, recall, and F\textsubscript{1}, treating one annotator as the reference. We adopt these metrics rather than Cohen's $\kappa$ \cite{Cohen:1968:weighted} to accommodate multiple errors and cross-dimensional explanations per word pair. Agreement is highest at the error type level (78.9--88.8 F\textsubscript{1}) and decreases at finer levels, reaching 68.6--75.6 F\textsubscript{1} at the explanation type level (Table~\ref{tab:iaa-agreement}).

\begin{table}[t]
\centering
\setlength{\tabcolsep}{3.8pt}
\begin{tabular}{llcccc}
\toprule
\textbf{Dimension} & \textbf{Type} & \textbf{Acc.} & \textbf{P} & \textbf{R} & \textbf{F\textsubscript{1}} \\
\midrule
\multirow{3}{*}{{Lex}} & Err. & 73.0 & 80.8 & 77.2 & 78.9 \\
 & Corr. & 71.0 & 76.4 & 83.5 & 78.3 \\
 & Exp. & 68.7 & 73.6 & 81.6 & 75.6 \\
\midrule
\multirow{3}{*}{{Orth}} & Err. & 83.8 & 87.7 & 90.6 & 88.8 \\
 & Corr. & 80.6 & 74.7 & 72.7 & 71.4 \\
 & Exp. & 79.1 & 71.1 & 74.0 & 70.8 \\
\midrule
\multirow{3}{*}{{Morph}} & Err. & 89.1 & 85.5 & 78.3 & 80.0 \\
 & Corr. & 86.7 & 72.5 & 74.2 & 72.2 \\
 & Exp. & 84.6 & 70.5 & 71.8 & 69.7 \\
\midrule
\multirow{3}{*}{{Synt}} & Err. & 84.2 & 83.3 & 88.9 & 83.5 \\
 & Corr. & 81.8 & 88.0 & 82.4 & 82.6 \\
 & Exp. & 71.9 & 70.9 & 72.2 & 68.6 \\
\bottomrule
\end{tabular}
\caption{Inter-annotator agreement across all 1,000 word pairs in terms of accuracy (Acc.) and macro-averaged precision (P), recall (R), and F\textsubscript{1}.}
\label{tab:iaa-agreement}
\end{table}






\subsection{Data Statistics}
Table~\ref{tab:gold-coverage-by-source} summarizes the distribution of annotated explanation paths and groups across datasets and linguistic dimensions. Across the 1,000 sampled word pairs, the annotation produced 1,439 explanation paths spanning 26 error types, 102 correction types, and 179 explanation types, out of 28, 177, and 374 defined in the taxonomy (Table~\ref{tab:taxonomy-stats}), respectively. Coverage decreases from coarser to finer levels, reflecting the increasing granularity of the taxonomy. Orthographic explanation paths account for the largest share across all three datasets.

The 1,439 explanation paths constitute 1,232 explanation groups, spanning 26 distinct combinations of error types, 98 of correction types, and 160 of explanation types, out of 43, 355, and 2,218 supported by the taxonomy, respectively. Of these groups, 207 (17\%) are cross-dimensional, with morphological--syntactic dependencies accounting for 78\% and orthographic--morphological dependencies for 22\%. No groups spanning three dimensions were observed. Appendix~\ref{app:detailed-stats} reports per-dataset taxonomy coverage, while Appendix~\ref{app:taxonomy-examples} presents explanation path frequencies and annotation examples across linguistic dimensions.

\begin{table*}[t]
\centering
\setlength{\tabcolsep}{4pt}
\begin{tabular}{lcccc|ccc}
\toprule
 & \multicolumn{4}{c}{} 
 & \multicolumn{3}{c}{\textbf{Unique}} \\
\cmidrule(lr){6-8}
\textbf{Dimension(s)} 
 & \textbf{QALB-14} & \textbf{QALB-15} & \textbf{ZAEBUC} 
 & \textbf{Total} & \textbf{Err.} & \textbf{Corr.} & \textbf{Exp.} \\
\midrule
\rowcolor{gray!15}
\multicolumn{8}{c}{\textit{Explanation Paths}} \\
Lex   & 125 & 93  & 13 & 231 & 2  & 6  & 10 \\
Orth  & 469 & 168 & 42 & 679 & 4	 & 31 & 	44 \\
Morph & 198 & 110 & 25 & 333 & 13 & 	48 & 	61 \\
Synt  & 118 & 58  & 20 & 196 & 7	 & 17	 & 64 \\
\textbf{Total}
 & \textbf{910} & \textbf{429} & \textbf{100} 
 & \textbf{1,439} & \textbf{26} & \textbf{102} & \textbf{179} \\
\midrule
\rowcolor{gray!15}
\multicolumn{8}{c}{\textit{Explanation Groups}} \\
Lex        & 125 & 93  & 13 & 231 & 2  & 6  & 10 \\
Orth       & 434 & 159 & 40 & 633 & 4	 & 26	 & 38 \\
Morph      & 62  & 57  & 7  & 126 & 6 & 	15 & 	25 \\
Synt       & 17  & 14  & 4  & 35  & 4 & 	9	 & 14 \\
Orth+Morph & 35  & 9   & 2  & 46  & 3	 & 7	 & 7 \\
Morph+Synt & 101 & 44  & 16 & 161 & 7	 & 35	 & 66\\
\textbf{Total}
 & \textbf{774} & \textbf{376} & \textbf{82} 
 & \textbf{1,232} & \textbf{26} & \textbf{98} & \textbf{160} \\
\bottomrule
\end{tabular}
\caption{Distribution of annotated explanation paths and groups across datasets and linguistic dimensions. Path counts include paths from both single- and cross-dimensional groups. Unique reports the number of distinct error (Err.), correction (Corr.), and explanation (Exp.) types for paths, and distinct combinations of these types for groups.}
\label{tab:gold-coverage-by-source}
\end{table*}

\section{Experimental Setup}
\label{sec:experiments}
\subsection{Task Formulation}
\label{sec:task-formulation}
Given an erroneous (source) sentence $s^{src}$, its corrected (target) counterpart $s^{tgt}$, and the aligned erroneous-corrected word pairs $A = \{(w_1^{src}, w_1^{tgt}), \dots, (w_n^{src}, w_n^{tgt})\}$ between them, obtained from existing word-level alignments in our source corpora (\S\ref{sec:data-selection}), we define GEE as a classification task over word pairs.\footnote{Unlike prior GEE work \cite{song-etal-2024-gee}, we isolate explanation from error identification using word alignment.}

Formally, let $\mathcal{P}$ denote the set of valid explanation paths defined by {\ArabiGEE}, where each path $p = (e, c, x) \in \mathcal{P}$ consists of an error type $e$, a correction type $c$, and an explanation type $x$ (\S\ref{sec:arabigee-design}). For each aligned pair $(w_i^{src}, w_i^{tgt}) \in A$, given the full sentence pair $(s^{src}, s^{tgt})$ as context, the task is to predict a set of explanation groups $G_i = \{g_{i1}, \dots, g_{ik}\}$, where $k \geq 1$ is the number of distinct errors in the pair, and each group $g_{ij} \subseteq \mathcal{P}$ is a non-empty set of paths jointly describing error $j$, possibly spanning multiple linguistic dimensions (\S\ref{sec:arabigee-design}). Each predicted path deterministically retrieves its explanation text, so no text is itself predicted or generated.

We evaluate LLMs on this task using structured prompting, since they require no task-specific training data to apply the taxonomy. Because prediction is grounded in the fixed, discrete space $\mathcal{P}$ rather than free-form text, our GEE formulation is model-agnostic: any predictor capable of scoring or selecting explanation groups, not only LLMs, can be evaluated under this task.





\paragraph{Input Variants} We test three input variants to study how LLMs use contextual and punctuation information when predicting explanation groups. In all settings, models see the erroneous-corrected word pair and at least the source sentence context. The \textbf{Full} variant includes both erroneous source and corrected target contexts with punctuation.  Since Arabic punctuation is often inconsistent and noisy \cite{alhafni-etal-2023-advancements,alhafni-habash-2025-enhancing,elkholy-etal-2026-arabic-sentence}, the \textbf{NoPnx} variant removes punctuation to test reliance on punctuation cues. Finally, the \textbf{NoTgt} variant removes the target context, to test whether models can predict accurate explanations from source context alone.

\subsection{LLMs}
We evaluate seven LLMs spanning proprietary and open-weight models. The proprietary models include GPT-5.5 \cite{gpt5}, Gemini-3.1-pro-preview \cite{gemini}, Qwen-3.5-plus \cite{qwen2025qwen25technicalreport}, and Claude-Opus-4.7 \cite{anthropic2025claudeopus4}. The open-weight models include Mistral-Large-3 \cite{mistralai2025large3}, DeepSeek-v4-pro \cite{deepseekai2026deepseekv4highlyefficientmilliontoken}, and the Arabic-centric Fanar-2-27B-Instruct \cite{fanar2}.



\begin{table*}[t]
\centering
\setlength{\tabcolsep}{4pt}
\begin{tabular}{l ccc ccc ccc}
\toprule
 & \multicolumn{3}{c}{\textbf{Full Acc.}} & \multicolumn{3}{c}{\textbf{Partial Acc.}} & \multicolumn{3}{c}{\textbf{Well-Formedness}} \\
\cmidrule(lr){2-4} \cmidrule(lr){5-7} \cmidrule(lr){8-10}
\textbf{Model} & \textbf{Full} & \textbf{NoTgt} & \textbf{NoPnx} & \textbf{Full} & \textbf{NoTgt} & \textbf{NoPnx} & \textbf{Full} & \textbf{NoTgt} & \textbf{NoPnx} \\
\midrule
GPT       & 82.3 & 81.5 & \underline{83.1} & 87.5 & 86.6 & 87.3 & \textbf{100.0} & \textbf{100.0} & \textbf{100.0} \\
Gemini    & \textbf{83.4} & \textbf{82.8} & \underline{\textbf{83.5}} & \textbf{87.8} & \textbf{87.5} & \textbf{87.8} & 99.7 & \textbf{100.0} & 99.8 \\
Opus      & 53.1 & 51.9 & \underline{53.6} & 59.9 & 59.4 & 60.8 & 96.4 & 95.7 & 96.9 \\
Qwen      & 56.2 & \underline{57.0} & 54.6 & 63.2 & 63.4 & 62.7 & 99.2 & 99.5 & 99.2 \\
DeepSeek  & 69.1 & \underline{69.5} & 67.9 & 75.9 & 76.7 & 75.9 & 99.0 & 98.1 & 98.9 \\
Mistral   & 21.1 & 21.9 & \underline{23.1} & 27.9 & 28.0 & 28.9 & 76.8 & 75.4 & 78.2 \\
Fanar     & 9.0 & \underline{11.3} & 10.4 & 11.2 & 13.3 & 12.7 & 82.7 & 85.7 & 83.7 \\
\bottomrule
\end{tabular}
\caption{Overall results across the three input variants. \textbf{Bold} indicates the best result in each column; \underline{underlining} marks the input variant in which a model achieves its highest Full Accuracy.}
\label{tab:overall}
\end{table*}


\paragraph{Prompting Strategy}
We use structured prompting based on {\ArabiGEE}. The prompt includes a compact taxonomy representation: for each explanation path, we provide its linguistic dimension, error type, explanation text, illustrative example, and cross-dimensional links where applicable. We omit higher-level taxonomy metadata to reduce prompt length, as much of this information is implicitly encoded in the explanation text itself. Models are instructed to predict an explanation group for each distinct error, respecting the taxonomy's structural and linguistic constraints. Due to Fanar's shorter context window, it receives a compact prompt omitting illustrative examples. The prompts are in Appendix~\ref{app:prompts}.


\paragraph{Output Matching} Since models may predict explanation groups in a different order from the gold annotations, we apply bipartite matching to identify the optimal alignment between predicted and reference groups before evaluation. The matching score is computed hierarchically over the taxonomy structure, allowing partial credit for predictions that agree at coarser taxonomy levels while differing at finer ones. Full details are provided in Appendix~\ref{app:matching}.




\subsection{Evaluation Metrics}
We evaluate predictions at three levels: word pairs, explanation groups, and explanation paths. At the word-pair level, \textbf{Full Accuracy} measures the percentage of word pairs whose predictions exactly match gold annotations, while \textbf{Partial Accuracy} measures the mean proportion of correctly predicted gold groups per word pair. At the group level, \textbf{Recall} measures the proportion of correctly predicted groups, while \textbf{Well-Formedness} measures the percentage of predicted groups satisfying {\ArabiGEE}'s structural and linguistic constraints (no out-of-taxonomy paths, duplicate dimensions, or incompatible combinations). At the path level, we report macro-averaged \textbf{precision}, \textbf{recall}, and \textbf{F\textsubscript{1}} per taxonomy level and linguistic dimension.

\section{Results}
\label{sec:results}

\subsection{Overall Performance}
Table~\ref{tab:overall} reports Full Accuracy, Partial Accuracy, and Well-Formedness across the three input variants. GPT and Gemini consistently achieve the strongest results, substantially outperforming the remaining models, with no statistically significant difference between the two in any setting.\footnote{Statistical significance was done using a paired approximate randomization test \cite{dror-etal-2018-hitchhikers}.} In contrast, Mistral and Fanar perform poorly across all metrics including Well-Formedness. Interestingly, Qwen and DeepSeek achieve very high Well-Formedness despite substantially lower accuracy, suggesting that structurally valid outputs are easier to produce than correct explanation groups.

\paragraph{Effect of Input Variants} The effect of input configuration varies substantially across models. GPT, Gemini, Opus, and Mistral achieve their best Full Accuracy under the NoPnx setting, suggesting punctuation may introduce noise. By contrast, Qwen, DeepSeek, and Fanar perform best under NoTgt, suggesting that the corrected target context does not always help. Despite these differences, the gaps between input variants remain small, within 1--2 points of Full Accuracy for the top-performing models. Given the inconsistent, noisy nature of Arabic punctuation (\S\ref{sec:arabic-facts}), punctuation appears to contribute only marginally once sufficient context is available. Appendix~\ref{app:per-dataset-overall-results} reports results across input variants for each of the three GEC datasets.

\subsection{Performance Across Explanation Groups}
\begin{table*}[t]
\centering
\begin{tabular}{lccc|ccc}
\toprule
 & \multicolumn{3}{c}{\textbf{GPT}}
 & \multicolumn{3}{c}{\textbf{Gemini}} \\
\cmidrule(lr){2-4}
\cmidrule(lr){5-7}
\textbf{Dimension(s)}
 & \textbf{Full} & \textbf{NoTgt} & \textbf{NoPnx}
 & \textbf{Full} & \textbf{NoTgt} & \textbf{NoPnx} \\
\midrule
\rowcolor{gray!15}
\multicolumn{7}{c}{\textit{Single-dimensional}} \\
Lex
 & 83.5 & \underline{84.4} & 83.1
 & \textbf{86.1} & \underline{\textbf{86.6}} & \textbf{85.3} \\
Orth
 & \textbf{87.4} & \textbf{87.8} & \underline{\textbf{88.2}}
 & 86.7 & 86.6 & \underline{87.2} \\
Morph
 & \underline{\textbf{93.7}} & \textbf{92.1} & \textbf{92.9}
 & 88.1 & \underline{91.3} & \underline{91.3} \\
Synt
 & \underline{48.6} & 42.9 & \textbf{45.7}
 & \underline{\textbf{51.4}} & \underline{\textbf{51.4}} & 42.9 \\
\textbf{Overall}
 & \textbf{86.0} & \textbf{86.0} & \underline{\textbf{86.1}}
 & 85.6 & \underline{\textbf{86.0}} & 85.8 \\
\midrule
\rowcolor{gray!15}
\multicolumn{7}{c}{\textit{Cross-dimensional}} \\
Orth+Morph
 & 95.7 & 95.7 & \underline{\textbf{97.8}}
 & \underline{\textbf{100.0}} & \underline{\textbf{100.0}} & \textbf{97.8} \\
Morph+Synt
 & \underline{\textbf{91.3}} & 83.9 & 89.4
 & 89.4 & \textbf{85.7} & \underline{\textbf{90.1}} \\
\textbf{Overall}
 & \underline{\textbf{92.3}} & 86.5 & 91.3
 & \underline{91.8} & \textbf{88.9} & \underline{\textbf{91.8}} \\
\bottomrule
\end{tabular}
\caption{Explanation group Recall across single- and cross-dimensional explanation groups for GPT and Gemini under each input variant.
\textbf{Bold} indicates the best result across models for each
input variant; \underline{underlining} marks the best input
variant for each model within each row.}
\label{tab:group-performance}
\end{table*}
Table~\ref{tab:group-performance} compares explanation group Recall for the two best-performing models, GPT and Gemini. Both models perform better on cross-dimensional than single-dimensional groups across all input variants. Orthographic--morphological groups achieve the highest recall (95.7--100.0\%), while single-dimensional syntactic groups are the most challenging (42.9--51.4\%). Model differences are statistically significant only for single-dimensional morphological groups under Full ($p<0.05$).

Across input variants, Recall on single-dimensional groups is largely stable, with differences not exceeding 0.5 points for either model. Cross-dimensional groups exhibit greater sensitivity to target context, with overall recall dropping by 5.8 and 2.9 points for GPT and Gemini, respectively, under NoTgt. This is primarily driven by morphological--syntactic groups, where recall drops by 7.4 and 3.7 points, respectively, highlighting the importance of target context in modeling morphosyntactic dependencies. Removing punctuation has minimal impact on overall group performance, consistent with our earlier findings.

\subsection{Performance Across Linguistic Dimensions}
\begin{table}[t]
\centering
\setlength{\tabcolsep}{6pt}
\begin{tabular}{l l cc}
\toprule
\textbf{Dimension} & \textbf{Setting} & \textbf{GPT} & \textbf{Gemini} \\
\midrule
\multirow{3}{*}{Lex} & Full & 77.0 & \textbf{82.3} \\
                        & NoTgt    & \textbf{78.4} & 80.7 \\
                        & NoPnx     & 76.9 & 80.4 \\
\midrule
\multirow{3}{*}{Orth}   & Full & \textbf{77.5} & \textbf{77.3} \\
                        & NoTgt    & 77.4 & 71.5 \\
                        & NoPnx    & 76.3 & 76.1 \\
\midrule
\multirow{3}{*}{Morph}  & Full & 80.8 & 80.2 \\
                        & NoTgt    & 77.8 & \textbf{82.6} \\
                        & NoPnx     & \textbf{81.6} & 79.0 \\
\midrule
\multirow{3}{*}{Synt} & Full & \textbf{79.0} & \textbf{84.1} \\
                        & NoTgt     & 70.4 & 77.6 \\
                        & NoPnx     & 74.7 & 82.9 \\

\bottomrule
\end{tabular}
\caption{Explanation path F\textsubscript{1} for GPT and Gemini across the three experimental setups. \textbf{Bold} indicates the best setting for each model per dimension.}
\label{tab:top2}
\end{table}
Table~\ref{tab:top2} compares GPT and Gemini at the explanation path level across linguistic dimensions and input variants. Orthography is the most challenging dimension overall, while both models perform strongly on morphology. Gemini consistently outperforms GPT on lexical and syntactic explanations, with statistically significant differences under Full for lexical and NoTgt for syntactic explanations ($p<0.05$).

Consistent with our group-level findings, the effect of input variants is generally limited, with syntax being particularly sensitive to the removal of target context. Under NoTgt, syntactic F\textsubscript{1} drops by 8.6 and 6.5 points for GPT and Gemini, respectively, further highlighting the importance of target context for syntactic explanations. Removing punctuation also affects syntactic explanations, particularly for GPT (4.3 drop). Detailed results across models, taxonomy levels, and datasets are provided in Appendices~\ref{app:detailed-results} and~\ref{app:detailed-full-variant-results-dataset}.

Appendix~\ref{sec:error-analysis} provides a detailed error analysis at both the explanation group and path levels.

\section{Conclusion and Future Work}
We introduced {\ArabiGEE}, a hierarchical taxonomy for Arabic GEE spanning lexical, orthographic, morphological, and syntactic dimensions, comprising 28 error types, 177 correction types, and 374 explanation types, supporting 2,218 explanation groups. Unlike prior Arabic error annotation frameworks for flat error typing, {\ArabiGEE} captures cross-dimensional linguistic dependencies and enables formulating GEE as a structured prediction task with automatic evaluation. We applied {\ArabiGEE} to manually annotate 1,000 word pairs from three Arabic GEC corpora, producing 1,232 groups comprising 1,439 explanation paths. We demonstrated the usefulness of our framework by evaluating a diverse set of LLMs on Arabic GEE.


In future work, we plan to expand the annotations, develop error type annotation tools based on {\ArabiGEE}, build models that predict explanations directly, and validate the taxonomy's pedagogical effectiveness with Arabic learners.



\section*{Limitations}
Our annotated set covers only 1,000 word pairs and may not capture the full distribution of Arabic learner errors. The explanations were designed by experts but have not yet been evaluated with Arabic learners for pedagogical effectiveness. Our taxonomy does not model pragmatic-level errors, such as discourse-level appropriateness or register-sensitive choices; extending it to this level is left to future work. Moreover, our LLM evaluation covers only one Arabic-centric model; evaluating more Arabic-centric models is left to future work.



\section*{Ethics Statement}
This work aims to support Arabic language learning through structured grammatical explanations. The study uses existing Arabic GEC corpora and does not introduce new personally identifying information. Automatically predicted explanations may be incorrect or overly prescriptive, especially given Arabic dialectal variation and the focus on Modern Standard Arabic. Such systems should therefore be used as assistive tools rather than replacements for teachers.

We used AI writing assistance within the scope of ``Assistance purely with the language of the paper'' described in the ACL Policy on Publication Ethics.

\section*{Acknowledgments}
This work was supported by the MBZUAI Arabic Language and Culture (MALAC) Center at the Mohamed bin Zayed University of Artificial Intelligence (MBZUAI), with additional support from Google.org and the Google Cloud Research Credits program through the Gemini Academic Program. We thank Hanada Taha for insightful discussions.

\bibliography{custom,anthology-1,anthology-2,camel-bib-v3}

\appendix

\newpage

\section{License and Cost}
\label{app:license}
We list below the licenses of the data and tools used in this work, all of which are employed in accordance with their intended use.

\begin{itemize}
    \item QALB-2014~\cite{mohit-etal-2014-first}: MIT License.
    \item QALB-2015~\cite{rozovskaya-etal-2015-second}: MIT License.
    \item ZAEBUC~\cite{habash-palfreyman-2022-zaebuc}: MIT License.
    \item GPT-5.5~\cite{gpt5}: Proprietary API model accessed via OpenAI API.
   \item Gemini-3.1-pro-preview~\cite{gemini}: Proprietary API model accessed via the Google Gemini API.
    \item Qwen-3.5-plus~\cite{qwen2025qwen25technicalreport}: Proprietary API model accessed via OpenRouter.
    \item Mistral-Large-3~\cite{mistralai2025large3}: open-weight model released under the Apache 2.0 license; accessed via OpenRouter.
    \item DeepSeek-v4-pro~\cite{deepseekai2026deepseekv4highlyefficientmilliontoken}: open-weight model released under the MIT license; accessed via OpenRouter.
    \item Fanar-2-27B-Instruct~\cite{fanar2}: open-weight model released under the Apache 2.0 license; run locally with vLLM.
    \item Claude-Opus-4.7~\cite{anthropic2025claudeopus4}: Proprietary API model accessed via OpenRouter.
\end{itemize}
In terms of cost, we used \$1,055 in credits to prompt the commercial LLMs via OpenRouter.

\section{GEC Dataset Statistics}
\label{app:app-gec-data-stats}
\begin{table}[ht]
\centering
\small
\setlength{\tabcolsep}{3pt}
\begin{tabular}{llcccl}
\toprule
\bf Dataset &  \bf Split & \multicolumn{1}{c}{\bf Words} & \bf Err.  & \bf Type & \bf Domain \\\hline
\multirow{3}{*}{\bf QALB-2014} & Train-L1  & 1M & 30\% & L1 & Comments \\
                               & Dev-L1  & 54K & 31\% & L1 & Comments \\
                               & Test-L1 & 51K &  32\% & L1 & Comments \\\hline

\multirow{3}{*}{\bf QALB-2015} & Train-L2  & 43K & 30\% & L2 & Essays \\
                               & Dev-L2  & 25K & 29\% & L2 & Essays \\
                               & Test-L2  & 23K & 29\% & L2 & Essays \\\hline

\multirow{3}{*}{\bf ZAEBUC} & Train-L1  & 25K &  24\% & L1 & Essays \\
                            & Dev-L1 & 5K & 25\% & L1 & Essays \\
                            & Test-L1  & 5K & 26\% & L1 & Essays \\
\bottomrule

\end{tabular}
\caption{Corpus statistics of Arabic GEC datasets.}
\label{tab:data-stats}
\end{table}

\newpage

\section{Inter-annotator Disagreement Analysis}

\begin{table}[t]
\centering
\begin{tabular}{lcccc}
\toprule
 & & \multicolumn{3}{c}{\textbf{Type}} \\
\cmidrule(lr){3-5}
\textbf{Dimension} & \textbf{Detection} & \textbf{Err.} & \textbf{Corr.} & \textbf{Exp.} \\
\midrule
Lex       & 19.7 & 12.2 & 2.3 & 2.8  \\
Orth  & 36.2 & 9.4  & 9.9 & 4.7  \\
Morph & 16.9 & 0.0  & 3.8 & 3.3 \\
Synt     & 14.1 & 1.4  & 1.9 & 9.4 \\
\bottomrule
\end{tabular}
\caption{Inter-annotator disagreements across linguistic dimensions, as percentages of the 213 disagreeing word pairs.}
\label{tab:iaa-disagreement}
\end{table}

\label{app:iaa-disagreements}
We distinguish three types of disagreements: (1) \textit{error-count disagreements}, where annotators identify different numbers of explanation groups (i.e., numbers of errors); (2) \textit{dimension-detection disagreements}, where a dimension is present in only one of two matched groups or belongs to an unmatched group; and (3) \textit{type-assignment disagreements}, where annotations differ in their explanation paths within shared dimensions of matched groups. These categories are not mutually exclusive, as error-count disagreements also give rise to dimension-detection disagreements.

Table~\ref{tab:iaa-disagreement} presents the distribution of dimension-detection and type-assignment disagreements across the four linguistic dimensions among the 213 disagreeing word pairs.

\section{Prompts}
\label{app:prompts}
The prompts used to evaluate LLMs are below. The system prompt (Box~\ref{box:system-prompt}) establishes the task, the input/output format, and the taxonomy's four dimensions, and includes five few-shot examples covering distinct prediction patterns: combining orthographic and morphological dimensions for a single error; combining morphological and syntactic dimensions without the orthographic dimension; separating multiple distinct errors within one aligned pair; a purely lexical error; and a case where a specific morphological explanation is preferred over a generic orthographic one. The output format (Box~\ref{box:output-format}) is conveyed through these few-shot examples rather than stated explicitly. The user template (Box~\ref{box:user-prompt}) is instantiated per example with the taxonomy definitions for each dimension, the source and target contexts, and the aligned word pairs to be predicted on. For Fanar, we use a compact version of the prompt that omits the illustrative examples from the taxonomy.

\newcounter{box}

\phantomsection
\refstepcounter{box}
\label{box:system-prompt}

\begin{tcolorbox}[
  breakable,
  colback=white,
  colframe=orange!60!black,
  title=\textbf{System Prompt},
  fontupper=\small,
]
\ttfamily
\begin{verbatim}
You are an annotation assistant for Arabic
grammatical error explanation.

You receive:
- the erroneous Arabic context (the learner
  sentence)
- the corrected Arabic context (the target
  sentence)
- the erroneous aligned word pairs, each with
  `alignment_id`, `wrong`, and `correct`
- the taxonomies of four dimensions:
  `lexical`, `orth`, `morph`, and `synt`

Your task: for each aligned word pair, find
every distinct error and explain why it is
needed, using explanation codes from the
taxonomies.
- The taxonomy has four dimensions:
  - `lexical`, used for word choice, dialect,
    and missing or extra words and clitics.
  - `orth`, used for spelling: hamza seat,
    taa marbuta, spacing, letters, vowels,
    and punctuation.
  - `morph`, used for word form: case,
    gender, and number markers, state,
    pattern, and stem vowel.
  - `synt`, used for the syntactic reason
    that requires a form: subject, object,
    predicate, agreement, and dependency.
- Each dimension groups its explanations by
  error type. Output an explanation's `code`,
  and use its `explanation` and
  `example_pairs` to decide when it applies.
  Codes are local to their dimension: `ex3`
  in `orth` and `ex3` in `morph` are
  different explanations.

\end{verbatim}
\end{tcolorbox}
\newpage
\begin{tcolorbox}[
  breakable,
  colback=white,
  colframe=orange!60!black,
  fontupper=\small,
]
\ttfamily
\begin{verbatim}
- An empty `wrong` ("") means the learner
  omitted something; an empty `correct`
  means the learner added something that
  must be deleted.
- An error is one change from `wrong` to
  `correct`. Several dimensions can explain
  the same error, each as a layer: for
  example, a hamza seat error may need
  orth + morph, and a gender agreement error
  morph + synt.
- Link values connect dimensions:
  `morpho_orthographic` connects orth and
  morph, and `morpho_syntactic` connects
  morph and synt. Linked explanations can
  explain one error together only if their
  values are equal.

Each dimension taxonomy has this structure:
{
  "dimension": "orth",
  "error_types": [
    {
      "error_type": "Hamza",
      "explanations": [
        {
          "code": "ex3",
          "explanation": "...",
          "example_pairs": [
            {"wrong": "...", "correct": "...",
             "context": "..."}
          ],
          "morpho_orthographic": "Kasra"
        }
      ]
    }
  ]
}

Few-shot examples:
{few_shot_examples}

\end{verbatim}
\end{tcolorbox}


\phantomsection
\refstepcounter{box}
\label{box:user-prompt}
\begin{tcolorbox}[
  breakable,
  colback=white,
  colframe=orange!60!black,
  title=\textbf{User Prompt},
  fontupper=\small,
]
\ttfamily

\begin{verbatim}
Now annotate the following item.

Input:
- Erroneous context: the learner sentence.
- Corrected context: the target sentence.
- Erroneous aligned word pairs: only the
  pairs that contain errors, each with
  `alignment_id`, `wrong`, and `correct`.
- Dimension taxonomies: the only codes you
  may select.

Available lexical explanations:
{lexical_taxonomy_json}

Available orthographic explanations:
{orth_taxonomy_json}

Available morphological explanations:
{morph_taxonomy_json}

Available syntactic explanations:
{synt_taxonomy_json}

{source_context_section}
{corrected_context_section}

Erroneous aligned word pairs:
{alignments_json}

\end{verbatim}
\end{tcolorbox}

\begin{tcolorbox}[
  breakable,
  colback=white,
  colframe=orange!60!black,
  fontupper=\small,
]
\ttfamily

\begin{verbatim}
Rules:
1. Annotate every aligned pair exactly once,
   and keep its `alignment_id` unchanged.
2. Find every distinct error in a pair, and
   return each one as its own error object.
3. Together, the errors must account for
   every change from `wrong` to `correct`; no
   error may describe a change that did not
   happen.
4. For each error, include every dimension
   that explains it, and only those.
5. Use at most one explanation per dimension
   in an error.
6. A lexical explanation stands alone in its
   error.
7. Dimensions combine through morph: orth +
   morph and morph + synt are valid, and orth
   + synt is valid only with morph as well
   (orth + morph + synt).
8. Orthographic and morphological
   explanations can share an error only if
   their `morpho_orthographic` values are
   equal, and morphological and syntactic
   explanations only if their
   `morpho_syntactic` values are equal. An
   explanation without that value cannot be
   combined with that dimension. These values
   are not output fields.
9. In each dimension, choose the most
   specific explanation that fits. Use a
   generic explanation only when no specific
   one applies.
10. A hamza on the wrong seat gets a hamza
    explanation. A middle or final hamza that
    is not written at all is a letter
    substitution or insertion. A missing
    initial hamza gets a hamza explanation.
11. A letter replaced by a space is two
    errors: one for the letter and one for
    the spacing.
12. A swapped letter order is orthographic,
    not a lexical substitution.
13. If `wrong` is not a real word, prefer an
    orthographic explanation over a lexical
    one, unless it is a dialectal or foreign
    word.
14. A dialectal form gets the dialect
    explanation, in addition to any other
    error in the pair.
15. Use a lexical clitic explanation when a
    clitic is added, deleted, or replaced,
    and a lexical word explanation when a
    separate word is added or deleted. Spacing
    around a clitic is orthographic.
16. Use only the dimension names `lexical`,
    `orth`, `morph`, `synt`, and only codes
    from the given taxonomies.
17. Return only a JSON object in the same
    format as the few-shot outputs, with codes
    only: no explanation text, examples, error
    types, link values, or commentary.

\end{verbatim}
\end{tcolorbox}

\phantomsection
\refstepcounter{box}
\label{box:output-format}
\begin{tcolorbox}[
  colback=white,
  colframe=orange!60!black,
  title=\textbf{Output Format},
  fonttitle=\bfseries\normalsize,
  fontupper=\small,
  breakable,
]
\ttfamily

\begin{verbatim}
{
  "annotations": [
    {
      "alignment_id": 4,
      "errors": [
        {
          "explanations": [
            {"dimension": "morph", "code": "ex63"}
          ]
        },
        {
          "explanations": [
            {"dimension": "orth", "code": "ex59"}
          ]
        }
      ]
    }
  ]
}
\end{verbatim}
\end{tcolorbox}

\clearpage
\section{Matching Algorithm}
\label{app:matching}


When a single source-target pair contains multiple annotated errors, annotators and models may produce different numbers of explanation groups in different orders. To compare two sets of explanation groups independently of their ordering, we use a global assignment procedure that maximizes taxonomy-level agreement between the groups. We employ this procedure for both inter-annotator agreement and model evaluation. For simplicity, we describe the algorithm below in the context of model evaluation, using predicted and gold explanation groups.

\paragraph{Group Similarity}
Following \S\ref{sec:task-formulation}, let $G_i$ and $\hat{G}_i$ denote the gold and predicted explanation groups of word pair $i$. For a group $g$, let $g^d$ denote its explanation path in dimension $d$. We compare two paths level by level, from the error type through the correction type to the explanation type, and measure their agreement as the number of leading levels they share, $\text{LCP}(g^d, \hat{g}^d) \in \{0, 1, 2, 3\}$. A dimension present in only one of the two groups receives an agreement of $0$, and a dimension absent from both is ignored. We define the similarity between a gold group $g$ and a predicted group $\hat{g}$ as:
\begin{equation}
S(g, \hat{g}) = \frac{1}{3} \sum_{d} \text{LCP}(g^d, \hat{g}^d).
\end{equation}
This similarity gives partial credit to predictions that agree with the gold group at coarser taxonomy levels. For example, the gold group of Error \#2 in Figure~\ref{fig:intro-example} consists of three paths: \textit{Hamza} $\rightarrow$ \textit{Middle Waw} $\rightarrow$ \textit{Hamza With Damma} (orthographic), \textit{Case Marker} $\rightarrow$ \textit{Damma} $\rightarrow$ \textit{Broken Plural Nominative} (morphological), and \textit{Case} $\rightarrow$ \textit{Nominative} $\rightarrow$ \textit{Assign Subject} (syntactic). A predicted group with the same orthographic path, the morphological path \textit{Case Marker} $\rightarrow$ \textit{Damma} $\rightarrow$ \textit{Singular Nominative}, and no syntactic path receives a similarity of $(3 + 2 + 0)/3 \approx 1.67$.

\paragraph{Alignment Procedure}
We align the predicted and gold groups as follows:
\begin{enumerate}
    \item Within a single word pair, we compute the similarity $S(g, \hat{g})$ for every gold group $g \in G_i$ and predicted group $\hat{g} \in \hat{G}_i$. If a predicted group contains more than one path in the same dimension, we compare the gold path with the closest predicted path and treat the remaining paths as extra.
    \item Next, we select the one-to-one assignment between $G_i$ and $\hat{G}_i$ with the highest total similarity, enumerating all possible assignments since a word pair contains only a few groups.

    \item Finally, we count the gold groups left unmatched as \textit{missing} and the predicted groups left unmatched as \textit{extra}.
\end{enumerate}
For instance, if a word pair contains a lexical and a syntactic gold group, and the model predicts the same two groups in the opposite order, each predicted group is matched to its gold counterpart rather than to the gold group in the same position.

\subsection{Properties}
This procedure has three important properties:

\begin{itemize}
    \item \textbf{Order-invariant.} The matcher does not penalize the model for emitting explanations in a different order from the gold reference.
    \item \textbf{Partial-credit aware.} Because the scoring function compares hierarchical taxonomy paths via longest shared prefix, near-misses at finer taxonomy levels contribute proportionally to the alignment without being treated identically to complete mismatches.
    \item \textbf{Cardinality-tolerant.} The matcher handles cases where the model emits more or fewer groups than the gold reference, separating alignment quality from over- and under-prediction, which are tracked as \textit{extra} and \textit{missing} counts.
\end{itemize}



\newpage

\begin{table}[t]
\centering
\setlength{\tabcolsep}{1.8pt}
\begin{tabular}{lcc|cc}
\toprule
 & \multicolumn{2}{c}{\textbf{Matched}}
 & \multicolumn{2}{c}{\textbf{Unmatched}} \\
\cmidrule(lr){2-3}
\cmidrule(lr){4-5}
\textbf{Dimension(s)}
 & \textbf{Correct} & \textbf{Wrong}
 & \textbf{Extra} & \textbf{Missing} \\
\midrule
\rowcolor{gray!15}
\multicolumn{5}{c}{\textit{Single-dimensional}} \\
Lex   & 197 & 33 & 6  & 1  \\
Orth  & 552 & 64 & 14 & 17 \\
Morph & 115 & 9  & 1  & 2  \\
Synt  & 15  & 20 & 1  & 0  \\
\textbf{Subtotal} & \textbf{879} & \textbf{126} & \textbf{22} & \textbf{20} \\
\midrule
\rowcolor{gray!15}
\multicolumn{5}{c}{\textit{Cross-dimensional}} \\
Orth+Morph      & 45  & 1  & 3  & 0 \\
Morph+Synt      & 145 & 13 & 8  & 3 \\
\textbf{Subtotal} & \textbf{190} & \textbf{14} & \textbf{11} & \textbf{3} \\
\midrule
\textbf{Total} & \textbf{1,069} & \textbf{140} & \textbf{33} & \textbf{23} \\
\bottomrule
\end{tabular}
\caption{Error analysis of Gemini predictions at the
Explanation Group level.}
\label{tab:group-error-analysis}
\end{table}
\begin{table}[h]
\centering
\setlength{\tabcolsep}{3pt}
\begin{tabular}{l cccc}
\toprule
 & \multicolumn{2}{c}{\textbf{Matched}} & \multicolumn{2}{c}{\textbf{Unmatched}} \\
\cmidrule(lr){2-3} \cmidrule(lr){4-5}
\textbf{Dimension} & \textbf{Correct} & \textbf{Wrong} & \textbf{Extra} & \textbf{Missing} \\
\midrule
Lex & 197 & 22 & 38 & 12 \\
Orth   & 601 & 41 & 33 & 37 \\
Morph  & 315 &  4 & 31 & 14 \\
Synt & 163 & 12 & 13 & 21 \\
\textbf{Total} & \textbf{1,276} & \textbf{79}
               & \textbf{115} & \textbf{84} \\
\bottomrule
\end{tabular}
\caption{Error analysis of Gemini predictions at the Explanation Path level.}
\label{tab:error-analysis}
\end{table}

\section{Error Analysis}
\label{sec:error-analysis}
To better understand model failures, we analyze the predicted explanation groups and paths from our best-performing model (Gemini) under its best input setting (NoPnx). Following the matching algorithm in Appendix~\ref{app:matching}, we distinguish four categories: \textit{Matched Correct} (identical predicted and gold groups or paths), \textit{Matched Wrong} (matched but non-identical groups or paths), \textit{Unmatched Extra} (predictions without corresponding gold annotations), and \textit{Unmatched Missing} (gold annotations without corresponding predictions).

\paragraph{Explanation Groups} Table~\ref{tab:group-error-analysis} presents the distribution of errors across single- and cross-dimensional groups. Most errors involve incorrectly predicted matched groups (140), rather than extra (33) or missing (23) groups. This is particularly pronounced for single-dimensional syntactic groups, where 20 of the 35 gold groups are matched incorrectly. Cross-dimensional groups exhibit fewer errors relative to their frequency, consistent with our findings in Table~\ref{tab:group-performance}.

\paragraph{Explanation Paths} Table~\ref{tab:error-analysis} presents the distribution of errors across linguistic dimensions. Unlike at the group level, most errors involve extra or missing paths rather than incorrect matched paths. This is particularly pronounced in morphology, with 31 extra and 14 missing paths compared to only four incorrect matched paths. Among incorrectly matched syntactic paths, 11 out of 12 share the same Error Type and Correction Type as the gold reference, differing only at the Explanation Type level. This indicates that the model generally identifies the appropriate syntactic correction but struggles with finer-grained linguistic distinctions.

\clearpage
\onecolumn
\section{Per-Dataset Taxonomy Coverage}
\label{app:detailed-stats}
\begin{table*}[h]
\centering
\setlength{\tabcolsep}{4pt}
\begin{tabular}{lccc|ccc|ccc}
\toprule
 & \multicolumn{3}{c}{\textbf{QALB-2014}}
 & \multicolumn{3}{c}{\textbf{QALB-2015}}
 & \multicolumn{3}{c}{\textbf{ZAEBUC}} \\
\cmidrule(lr){2-4}
\cmidrule(lr){5-7}
\cmidrule(lr){8-10}
\textbf{Dimension(s)}
 & \textbf{Err.} & \textbf{Corr.} & \textbf{Exp.}
 & \textbf{Err.} & \textbf{Corr.} & \textbf{Exp.}
 & \textbf{Err.} & \textbf{Corr.} & \textbf{Exp.} \\
\midrule
\rowcolor{gray!15}
\multicolumn{10}{c}{\textit{Explanation Paths}} \\
Lex   & 2  & 6  & 10 & 2 & 6  & 9  & 2 & 4  & 6  \\
Orth  & 4  & 29 & 40 & 4 & 21 & 27 & 4 & 11 & 14 \\
Morph & 12 & 40 & 48 & 8 & 28 & 36 & 7 & 13 & 15 \\
Synt  & 7  & 15 & 52 & 5 & 10 & 31 & 6 & 10 & 15 \\
\midrule
\rowcolor{gray!15}
\multicolumn{10}{c}{\textit{Explanation Groups}} \\
Lex        & 2	& 6	& 10	& 2	& 6	& 9	& 2	& 4	& 6 \\
Orth       & 4	& 24	& 34	& 4	& 20	& 26	& 4	& 10	& 13 \\
Morph      & 5	& 10	& 16	& 4	& 9	& 16	& 3	& 5	& 5 \\
Synt       & 4	& 8	& 11	& 3	& 4	& 5	& 2	& 3	& 3 \\
Orth+Morph & 3	& 7	& 7	& 1	& 1	& 1	& 1	& 1	& 1 \\
Morph+Synt & 7	& 27	& 51	& 5	& 20	& 31	& 5	& 9	& 12\\
\bottomrule
\end{tabular}
\caption{Per-dataset taxonomy coverage across linguistic dimensions.
We report the number of distinct error (Err.), correction (Corr.),
and explanation (Exp.) types for paths, and distinct combinations
of these types for groups.}
\label{tab:detailed-coverage-by-source}
\end{table*}

\newpage
\section{Per-Dataset Overall Results}
\label{app:per-dataset-overall-results}
\begin{table*}[th!]
\centering
\begin{tabular}{l ccc ccc ccc}
\toprule
 & \multicolumn{3}{c}{\textbf{Full Acc.}} & \multicolumn{3}{c}{\textbf{Partial Acc.}} & \multicolumn{3}{c}{\textbf{Well-Formedness}} \\
\cmidrule(lr){2-4} \cmidrule(lr){5-7} \cmidrule(lr){8-10}
\textbf{Model} & \textbf{Full} & \textbf{NoTgt} & \textbf{NoPnx} & \textbf{Full} & \textbf{NoTgt} & \textbf{NoPnx} & \textbf{Full} & \textbf{NoTgt} & \textbf{NoPnx} \\
\midrule
\multicolumn{10}{c}{\textbf{QALB-2014}} \\
\midrule
GPT      & 84.1 & 83.8 & \underline{84.6} & 88.6 & 88.3 & 88.6 & \textbf{100.0} & \textbf{100.0} & \textbf{100.0} \\
Gemini   & \textbf{85.2} & \textbf{85.1} & \underline{\textbf{85.7}} & \textbf{89.6} & \textbf{89.4} & \textbf{90.1} & 99.5 & \textbf{100.0} & 99.9 \\
Opus     & 54.6 & 53.9 & \underline{55.5} & 61.9 & 62.3 & 63.2 & 96.1 & 95.5 & 96.7 \\
Qwen     & \underline{58.1} & 57.8 & 56.5 & 65.4 & 65.3 & 64.4 & 99.6 & 99.6 & 99.2 \\
DeepSeek & 67.3 & \underline{67.9} & 66.3 & 74.5 & 74.9 & 73.5 & 98.4 & 97.1 & 98.3 \\
Mistral  & 21.3 & 21.8 & \underline{22.8} & 28.6 & 28.2 & 29.4 & 75.7 & 74.0 & 77.4 \\
Fanar    & 10.0 & \underline{11.4} & \underline{11.4} & 12.5 & 13.6 & 14.0 & 82.8 & 84.7 & 83.2 \\
\midrule
\multicolumn{10}{c}{\textbf{QALB-2015}} \\
\midrule
GPT      & 78.9 & 77.6 & \underline{\textbf{80.5}} & 85.4 & 83.7 & \textbf{85.1} & \textbf{100.0} & \textbf{100.0} & \textbf{100.0} \\
Gemini   & \underline{\textbf{81.2}} & \textbf{78.6} & 79.5 & \textbf{85.4} & \textbf{84.1} & 83.8 & \textbf{100.0} & \textbf{100.0} & 99.5 \\
Opus     & \underline{49.7} & 47.7 & 49.4 & 55.6 & 54.0 & 55.3 & 97.0 & 96.2 & 97.8 \\
Qwen     & 53.9 & \underline{55.5} & 50.3 & 60.5 & 59.7 & 59.6 & 98.6 & 99.1 & 99.2 \\
DeepSeek & \underline{72.4} & 71.4 & 69.5 & 77.8 & 79.7 & 78.9 & \textbf{100.0} & 99.7 & \textbf{100.0} \\
Mistral  & 20.5 & 21.4 & \underline{23.4} & 26.5 & 27.9 & 27.8 & 77.3 & 77.1 & 78.1 \\
Fanar    & 6.8 & \underline{10.4} & 7.5 & 8.4 & 12.4 & 9.0 & 80.2 & 87.3 & 83.4 \\
\midrule
\multicolumn{10}{c}{\textbf{ZAEBUC}} \\
\midrule
GPT      & \underline{\textbf{81.2}} & 78.3 & \underline{\textbf{81.2}} & \textbf{86.2} & 84.1 & \textbf{85.5} & \textbf{100.0} & \textbf{100.0} & \textbf{100.0} \\
Gemini   & 76.8 & \underline{\textbf{81.2}} & \underline{\textbf{81.2}} & 82.6 & \textbf{85.5} & \textbf{85.5} & \textbf{100.0} & \textbf{100.0} & \textbf{100.0} \\
Opus     & \underline{55.1} & 52.2 & \underline{55.1} & 60.9 & 58.0 & 63.0 & 96.3 & 96.2 & 95.1 \\
Qwen     & 49.3 & \underline{56.5} & \underline{56.5} & 56.5 & 61.6 & 60.9 & 97.5 & \textbf{100.0} & \textbf{100.0} \\
DeepSeek & 71.0 & \underline{75.4} & \underline{75.4} & 79.7 & 79.7 & 84.1 & \textbf{100.0} & \textbf{100.0} & \textbf{100.0} \\
Mistral  & 21.7 & \underline{24.6} & \underline{24.6} & 29.0 & 27.5 & 29.0 & 85.4 & 82.1 & 86.9 \\
Fanar    & 10.1 & \underline{14.5} & \underline{14.5} & 12.3 & 15.2 & 16.7 & 92.8 & 87.0 & 89.9 \\

\bottomrule
\end{tabular}
\caption{Overall results across the three input variants, broken down by dataset. \textbf{Bold} indicates the best result in each column within each dataset block; \underline{underlining} marks the input variant in which a model achieves its highest Full Accuracy.}
\label{tab:overall-results-per-dataset}
\end{table*}

\newpage
\section{Detailed Per-Dimension Results}
\label{app:detailed-results}
\begin{table*}[h!]
\centering
\begin{tabular}{l ccc ccc ccc}
\toprule
 & \multicolumn{3}{c}{\textbf{Error Type}} & \multicolumn{3}{c}{\textbf{Correction Type}} & \multicolumn{3}{c}{\textbf{Explanation Type}} \\
\cmidrule(lr){2-4} \cmidrule(lr){5-7} \cmidrule(lr){8-10}
\textbf{Model} & P & R & F\textsubscript{1} & P & R & F\textsubscript{1} & P & R & F\textsubscript{1} \\
\midrule
\multicolumn{10}{c}{\textbf{Lexical}} \\
\midrule
GPT      & 76.1 & 86.6 & 80.6 & 74.9 & 88.8 & 80.7 & 72.2 & 85.5 & 77.0 \\
Gemini   & \textbf{81.8} & \textbf{88.6} & \textbf{85.0} & \textbf{81.8} & \textbf{89.5} & \textbf{85.2} & \textbf{79.2} & \textbf{87.7} & \textbf{82.3} \\
Opus     & 72.2 & 80.0 & 75.9 & 69.6 & 79.7 & 73.6 & 68.7 & 70.5 & 67.3 \\
Qwen     & 68.6 & 73.5 & 70.9 & 75.6 & 76.4 & 75.7 & 69.6 & 63.2 & 63.0 \\
DeepSeek & 74.7 & 84.6 & 79.2 & 73.8 & 82.0 & 77.3 & 70.6 & 79.9 & 73.8 \\
Mistral  & 47.1 & 60.8 & 53.1 & 52.7 & 59.1 & 51.8 & 44.9 & 55.0 & 45.3 \\
Fanar    & 36.2 & 26.6 & 30.4 & 35.1 & 20.7 & 25.0 & 19.9 & 12.2 & 13.6 \\
\midrule
\multicolumn{10}{c}{\textbf{Orthography}} \\
\midrule
GPT      & 84.0 & 85.6 & 84.6 & 80.0 & \textbf{84.6} & \textbf{79.1} & 78.8 & \textbf{80.0} & \textbf{77.5} \\
Gemini   & \textbf{88.3} & \textbf{87.7} & \textbf{87.7} & \textbf{82.3} & 81.1 & 78.1 & \textbf{81.1} & 78.9 & 77.3 \\
Opus     & 78.5 & 70.7 & 71.4 & 51.1 & 48.7 & 43.8 & 43.4 & 40.7 & 37.8 \\
Qwen     & 76.0 & 68.7 & 68.1 & 56.4 & 49.6 & 47.9 & 39.2 & 34.6 & 32.7 \\
DeepSeek & 64.1 & 63.3 & 63.4 & 56.4 & 59.7 & 54.8 & 53.8 & 54.5 & 50.9 \\
Mistral  & 46.8 & 55.1 & 48.6 & 24.0 & 24.3 & 20.5 & 18.3 & 17.9 & 14.9 \\
Fanar    & 36.9 & 33.9 & 32.4 & 9.9 & 6.4 & 5.1 & 6.2 & 3.8 & 3.0 \\
\midrule
\multicolumn{10}{c}{\textbf{Morphology}} \\
\midrule
GPT      & \textbf{80.5} & 93.0 & \textbf{83.9} & \textbf{77.8} & 79.7 & \textbf{77.6} & \textbf{80.9} & 82.7 & \textbf{80.8} \\
Gemini   & 78.0 & \textbf{94.4} & 82.7 & 76.3 & \textbf{81.3} & 77.0 & 80.5 & \textbf{83.4} & 80.2 \\
Opus     & 65.6 & 74.1 & 65.0 & 50.1 & 48.4 & 47.3 & 47.2 & 44.4 & 44.0 \\
Qwen     & 60.4 & 60.1 & 56.5 & 46.8 & 45.9 & 44.0 & 45.7 & 44.2 & 42.8 \\
DeepSeek & 61.7 & 72.3 & 65.1 & 52.7 & 56.1 & 52.8 & 51.8 & 56.4 & 52.1 \\
Mistral  & 26.4 & 37.9 & 29.5 & 14.6 & 14.7 & 11.6 & 9.9 & 10.5 & 8.1 \\
Fanar    & 20.1 & 20.1 & 15.2 & 2.9 & 5.4 & 3.1 & 2.5 & 4.3 & 2.6 \\
\midrule
\multicolumn{10}{c}{\textbf{Syntax}} \\
\midrule
GPT      & 86.6 & 82.5 & 83.4 & 82.9 & 81.8 & 81.5 & 79.4 & 81.0 & 79.0 \\
Gemini   & \textbf{92.1} & \textbf{88.3} & \textbf{88.7} & \textbf{86.8} & \textbf{84.3} & \textbf{84.2} & \textbf{85.4} & \textbf{86.2} & \textbf{84.1} \\
Opus     & 67.1 & 70.3 & 62.6 & 61.4 & 63.9 & 58.4 & 41.6 & 40.3 & 36.8 \\
Qwen     & 69.2 & 55.2 & 60.6 & 72.7 & 55.9 & 61.5 & 46.3 & 43.4 & 42.6 \\
DeepSeek & 69.3 & 75.7 & 71.6 & 68.3 & 73.9 & 68.8 & 67.6 & 68.1 & 65.5 \\
Mistral  & 31.9 & 52.3 & 38.5 & 23.0 & 33.0 & 24.6 & 11.4 & 13.4 & 10.4 \\
Fanar    & 33.5 & 12.2 & 15.9 & 12.1 & 6.8 & 8.0 & 1.6 & 1.5 & 1.2 \\

\bottomrule
\end{tabular}

\caption{Per-dimension precision, recall, and F\textsubscript{1} at the Error Type, Correction Type, and Explanation Type levels under the \textbf{Full} setting. \textbf{Bold} indicates the best score in each column.}
\label{tab:full-full-input}
\end{table*}

\begin{table*}[th!]
\centering
\begin{tabular}{l ccc ccc ccc}
\toprule
 & \multicolumn{3}{c}{\textbf{Error Type}} & \multicolumn{3}{c}{\textbf{Correction Type}} & \multicolumn{3}{c}{\textbf{Explanation Type}} \\
\cmidrule(lr){2-4} \cmidrule(lr){5-7} \cmidrule(lr){8-10}
\textbf{Model} & P & R & F\textsubscript{1} & P & R & F\textsubscript{1} & P & R & F\textsubscript{1} \\
\midrule
\multicolumn{10}{c}{\textbf{Lexical}} \\
\midrule
GPT      & 76.6 & 86.7 & 80.8 & 76.6 & 88.8 & 81.6 & 74.5 & 86.0 & 78.4 \\
Gemini   & \textbf{78.8} & \textbf{88.5} & \textbf{83.4} & \textbf{80.3} & \textbf{89.3} & \textbf{84.3} & \textbf{76.9} & \textbf{87.7} & \textbf{80.7} \\
Opus     & 69.0 & 79.4 & 73.8 & 66.9 & 79.9 & 72.1 & 67.3 & 71.0 & 66.8 \\
Qwen     & 68.1 & 70.1 & 69.0 & 75.2 & 71.7 & 73.1 & 67.0 & 58.3 & 59.1 \\
DeepSeek & 74.7 & 85.0 & 79.4 & 74.1 & 84.0 & 78.5 & 73.2 & 81.0 & 75.5 \\
Mistral  & 49.2 & 59.4 & 53.4 & 53.8 & 55.6 & 50.4 & 44.9 & 54.0 & 44.8 \\
Fanar    & 38.5 & 35.5 & 36.5 & 33.0 & 25.5 & 27.5 & 21.6 & 17.0 & 17.0 \\
\midrule
\multicolumn{10}{c}{\textbf{Orthography}} \\
\midrule
GPT      & 85.2 & \textbf{87.1} & 85.9 & \textbf{80.5} & \textbf{85.2} & \textbf{79.4} & \textbf{78.6} & \textbf{80.6} & \textbf{77.4} \\
Gemini   & \textbf{89.5} & 84.3 & \textbf{86.2} & 78.0 & 77.9 & 74.4 & 75.2 & 72.3 & 71.5 \\
Opus     & 80.4 & 68.6 & 69.6 & 49.3 & 47.1 & 41.2 & 40.7 & 37.1 & 34.2 \\
Qwen     & 78.7 & 69.7 & 71.1 & 49.0 & 46.2 & 44.7 & 38.1 & 34.9 & 32.7 \\
DeepSeek & 79.4 & 78.1 & 78.3 & 60.5 & 62.1 & 58.4 & 54.2 & 54.2 & 51.6 \\
Mistral  & 52.6 & 55.5 & 49.4 & 21.3 & 27.0 & 19.2 & 16.1 & 15.9 & 12.5 \\
Fanar    & 38.2 & 32.7 & 32.6 & 10.8 & 6.3 & 5.5 & 6.5 & 3.9 & 3.3 \\
\midrule
\multicolumn{10}{c}{\textbf{Morphology}} \\
\midrule
GPT      & 79.7 & 84.5 & 81.6 & 74.4 & 74.9 & 74.1 & 78.1 & 78.5 & 77.8 \\
Gemini   & \textbf{80.4} & \textbf{94.4} & \textbf{83.7} & \textbf{79.8} & \textbf{82.8} & \textbf{79.7} & \textbf{82.9} & \textbf{84.9} & \textbf{82.6} \\
Opus     & 57.8 & 68.3 & 58.6 & 46.8 & 46.4 & 44.5 & 37.4 & 37.3 & 35.9 \\
Qwen     & 54.9 & 53.4 & 52.7 & 46.7 & 42.5 & 42.6 & 46.6 & 43.3 & 43.0 \\
DeepSeek & 60.6 & 72.9 & 61.6 & 57.4 & 61.4 & 57.1 & 55.6 & 57.2 & 54.5 \\
Mistral  & 27.9 & 41.3 & 30.3 & 15.9 & 17.9 & 14.1 & 12.0 & 13.3 & 10.5 \\
Fanar    & 20.4 & 18.4 & 15.8 & 5.2 & 5.4 & 4.7 & 4.1 & 4.3 & 3.7 \\
\midrule
\multicolumn{10}{c}{\textbf{Syntax}} \\
\midrule
GPT      & 86.7 & 80.8 & 82.4 & 80.9 & 78.6 & 78.7 & 72.0 & 72.7 & 70.4 \\
Gemini   & \textbf{91.2} & \textbf{85.0} & \textbf{86.2} & \textbf{88.8} & \textbf{80.3} & \textbf{82.4} & \textbf{80.1} & \textbf{78.0} & \textbf{77.6} \\
Opus     & 65.7 & 71.8 & 65.5 & 60.0 & 63.1 & 58.4 & 40.9 & 37.2 & 34.4 \\
Qwen     & 88.8 & 57.2 & 66.6 & 77.9 & 55.2 & 62.4 & 50.0 & 40.2 & 41.3 \\
DeepSeek & 76.0 & 82.3 & 78.0 & 71.2 & 75.5 & 71.2 & 62.8 & 65.9 & 61.9 \\
Mistral  & 38.8 & 54.2 & 43.6 & 28.7 & 31.6 & 26.8 & 12.8 & 13.3 & 10.9 \\
Fanar    & 36.4 & 19.7 & 23.3 & 21.4 & 10.5 & 12.4 & 2.4 & 2.3 & 2.1 \\

\bottomrule
\end{tabular}
\caption{Per-dimension precision, recall, and F\textsubscript{1} at the Error Type, Correction Type, and Explanation Type levels under the \textbf{NoTgt} setting. \textbf{Bold} indicates the best score in each column.}
\label{tab:full-no-trgt}
\end{table*}

\begin{table*}[t]
\centering
\begin{tabular}{l ccc ccc ccc}
\toprule
 & \multicolumn{3}{c}{\textbf{Error Type}} & \multicolumn{3}{c}{\textbf{Correction Type}} & \multicolumn{3}{c}{\textbf{Explanation Type}} \\
\cmidrule(lr){2-4} \cmidrule(lr){5-7} \cmidrule(lr){8-10}
\textbf{Model} & P & R & F\textsubscript{1} & P & R & F\textsubscript{1} & P & R & F\textsubscript{1} \\
\midrule
\multicolumn{10}{c}{\textbf{Lexical}} \\
\midrule
GPT      & 77.0 & 85.6 & 80.4 & 77.1 & \textbf{88.8} & 81.8 & 72.9 & 84.9 & 76.9 \\
Gemini   & \textbf{79.6} & \textbf{88.1} & \textbf{83.6} & \textbf{79.5} & 88.2 & \textbf{83.5} & \textbf{76.6} & \textbf{86.6} & \textbf{80.4} \\
Opus     & 69.5 & 76.9 & 73.0 & 67.0 & 77.3 & 70.6 & 62.9 & 66.8 & 63.0 \\
Qwen     & 67.7 & 70.9 & 69.2 & 73.9 & 75.9 & 74.8 & 58.8 & 58.8 & 57.3 \\
DeepSeek & 73.8 & 83.5 & 78.2 & 76.0 & 85.7 & 79.9 & 72.7 & 78.2 & 74.2 \\
Mistral  & 49.8 & 61.8 & 55.2 & 49.9 & 58.9 & 51.2 & 45.8 & 54.4 & 46.4 \\
Fanar    & 35.2 & 24.3 & 28.6 & 34.5 & 18.8 & 23.5 & 18.3 & 11.7 & 13.2 \\
\midrule
\multicolumn{10}{c}{\textbf{Orthography}} \\
\midrule
GPT      & 86.5 & \textbf{88.2} & \textbf{87.2} & 79.0 & \textbf{80.6} & \textbf{77.3} & 78.3 & 77.5 & \textbf{76.3} \\
Gemini   & \textbf{89.5} & 85.0 & 86.7 & \textbf{80.1} & 80.5 & 77.2 & \textbf{78.7} & \textbf{78.0} & 76.1 \\
Opus     & 83.5 & 72.7 & 74.1 & 48.4 & 46.7 & 40.5 & 38.0 & 35.4 & 32.1 \\
Qwen     & 80.2 & 66.9 & 66.6 & 46.9 & 46.3 & 43.2 & 36.0 & 33.1 & 31.2 \\
DeepSeek & 76.2 & 76.8 & 75.9 & 53.2 & 58.5 & 52.8 & 51.5 & 53.9 & 49.7 \\
Mistral  & 52.4 & 55.0 & 49.7 & 28.4 & 26.9 & 23.0 & 22.6 & 20.2 & 17.6 \\
Fanar    & 36.2 & 33.2 & 31.4 & 9.7 & 6.7 & 5.5 & 5.3 & 4.0 & 3.1 \\
\midrule
\multicolumn{10}{c}{\textbf{Morphology}} \\
\midrule
GPT      & \textbf{80.8} & 87.4 & \textbf{83.4} & \textbf{79.0} & \textbf{80.5} & \textbf{79.1} & \textbf{81.5} & \textbf{82.9} & \textbf{81.6} \\
Gemini   & 77.3 & \textbf{93.5} & 82.1 & 75.5 & 77.8 & 76.0 & 78.7 & 80.8 & 79.0 \\
Opus     & 61.8 & 72.6 & 62.0 & 47.1 & 47.6 & 45.8 & 45.5 & 41.8 & 41.5 \\
Qwen     & 53.2 & 52.3 & 50.8 & 49.9 & 50.8 & 48.3 & 46.7 & 47.1 & 44.9 \\
DeepSeek & 53.0 & 65.0 & 56.9 & 50.7 & 56.1 & 51.4 & 52.4 & 57.8 & 53.2 \\
Mistral  & 29.6 & 42.3 & 32.5 & 13.5 & 13.7 & 11.4 & 11.3 & 10.4 & 8.9 \\
Fanar    & 22.7 & 21.5 & 16.6 & 3.2 & 5.2 & 3.0 & 2.8 & 4.2 & 2.5 \\
\midrule
\multicolumn{10}{c}{\textbf{Syntax}} \\
\midrule
GPT      & 89.4 & 82.8 & 85.1 & 85.6 & 79.0 & 81.4 & 74.9 & 77.2 & 74.7 \\
Gemini   & \textbf{91.3} & \textbf{86.9} & \textbf{87.5} & \textbf{86.8} & \textbf{81.9} & \textbf{82.9} & \textbf{83.5} & \textbf{84.4} & \textbf{82.9} \\
Opus     & 69.2 & 76.4 & 68.5 & 59.3 & 64.4 & 57.8 & 43.4 & 41.4 & 38.7 \\
Qwen     & 58.4 & 46.2 & 50.5 & 60.9 & 48.3 & 52.3 & 45.9 & 40.9 & 41.2 \\
DeepSeek & 72.9 & 71.6 & 71.6 & 69.2 & 68.0 & 67.4 & 65.9 & 64.3 & 62.3 \\
Mistral  & 32.4 & 49.0 & 37.3 & 24.8 & 33.0 & 26.0 & 13.2 & 17.2 & 12.3 \\
Fanar    & 38.3 & 14.6 & 19.0 & 18.9 & 9.7 & 11.6 & 4.1 & 3.0 & 2.8 \\

\bottomrule
\end{tabular}
\caption{Per-dimension precision, recall, and F\textsubscript{1} at the Error Type, Correction Type, and Explanation Type levels under the \textbf{NoPnx} setting. \textbf{Bold} indicates the best score in each column.}
\label{tab:full-no-pnx}
\end{table*}

\newpage
\clearpage
\section{Per-Dimension Results by Dataset (Full Setting)}
\label{app:detailed-full-variant-results-dataset}
\begin{table}[th!]
\centering
\begin{tabular}{l ccc ccc ccc}
\toprule
& \multicolumn{3}{c}{\textbf{Error Type}} & \multicolumn{3}{c}{\textbf{Correction Type}} & \multicolumn{3}{c}{\textbf{Explanation Type}} \\
\cmidrule(lr){2-4} \cmidrule(lr){5-7} \cmidrule(lr){8-10}
\textbf{Model} & P & R & F\textsubscript{1} & P & R & F\textsubscript{1} & P & R & F\textsubscript{1} \\
\midrule
\multicolumn{10}{c}{\textbf{Lexical}} \\
\midrule
GPT      & 77.5 & 86.7 & 81.2 & 77.8 & 90.2 & 82.5 & 72.2 & 87.7 & 77.6 \\
Gemini   & \textbf{87.0} & \textbf{89.0} & \textbf{87.7} & \textbf{86.8} & \textbf{92.9} & \textbf{89.3} & \textbf{82.8} & \textbf{91.2} & \textbf{85.1} \\
Opus     & 76.9 & 81.1 & 78.7 & 79.3 & 88.1 & 82.9 & 74.8 & 76.2 & 72.7 \\
Qwen     & 71.5 & 72.9 & 71.9 & 80.1 & 83.0 & 80.9 & 69.7 & 67.9 & 65.5 \\
DeepSeek & 72.9 & 81.8 & 76.8 & 73.4 & 82.7 & 76.7 & 67.6 & 79.9 & 71.8 \\
Mistral  & 40.9 & 59.3 & 48.3 & 44.9 & 60.6 & 47.9 & 38.7 & 55.9 & 41.3 \\
Fanar    & 36.2 & 28.5 & 31.8 & 37.3 & 22.2 & 25.6 & 20.5 & 13.1 & 14.7 \\
\midrule
\multicolumn{10}{c}{\textbf{Orthography}} \\
\midrule
GPT      & 81.4 & 86.0 & 83.6 & 75.8 & 78.6 & 75.0 & \textbf{79.1} & \textbf{80.8} & \textbf{78.1} \\
Gemini   & \textbf{86.9} & \textbf{88.2} & \textbf{87.3} & \textbf{78.0} & \textbf{78.9} & \textbf{75.3} & 79.0 & 78.2 & 76.2 \\
Opus     & 78.0 & 70.5 & 71.1 & 48.0 & 49.0 & 43.8 & 42.7 & 41.8 & 38.5 \\
Qwen     & 75.5 & 69.1 & 68.9 & 59.1 & 55.5 & 53.3 & 42.1 & 39.1 & 37.0 \\
DeepSeek & 78.5 & 79.2 & 78.7 & 60.0 & 60.4 & 57.3 & 58.5 & 58.1 & 55.0 \\
Mistral  & 48.5 & 54.8 & 49.5 & 26.3 & 26.3 & 22.4 & 19.9 & 19.5 & 16.5 \\
Fanar    & 38.3 & 33.3 & 33.2 & 11.5 & 7.3 & 5.4 & 8.0 & 4.0 & 3.3 \\
\midrule
\multicolumn{10}{c}{\textbf{Morphology}} \\
\midrule
GPT      & 83.7 & 89.8 & 85.9 & \textbf{84.4} & 85.3 & \textbf{84.1} & \textbf{85.2} & 86.3 & \textbf{85.0} \\
Gemini   & \textbf{87.5} & \textbf{94.0} & \textbf{89.4} & 81.7 & \textbf{88.1} & 83.8 & 82.1 & \textbf{87.7} & 84.0 \\
Opus     & 69.7 & 74.6 & 68.3 & 52.8 & 54.7 & 52.4 & 52.4 & 50.4 & 49.8 \\
Qwen     & 60.6 & 58.3 & 57.3 & 54.5 & 54.6 & 52.4 & 51.9 & 50.7 & 48.9 \\
DeepSeek & 52.4 & 62.2 & 55.5 & 46.6 & 53.0 & 47.8 & 43.5 & 49.0 & 44.4 \\
Mistral  & 22.8 & 32.2 & 25.1 & 14.5 & 16.1 & 12.1 & 9.3 & 10.2 & 7.9 \\
Fanar    & 17.5 & 14.9 & 12.7 & 2.6 & 6.2 & 3.2 & 2.4 & 5.2 & 2.8 \\
\midrule
\multicolumn{10}{c}{\textbf{Syntax}} \\
\midrule
GPT      & 88.6 & 83.3 & 84.7 & \textbf{87.3} & 85.1 & 85.6 & 80.8 & 80.4 & 79.6 \\
Gemini   & \textbf{89.2} & \textbf{87.7} & \textbf{87.8} & 86.5 & \textbf{86.2} & \textbf{86.2} & \textbf{85.4} & \textbf{86.9} & \textbf{85.1} \\
Opus     & 69.9 & 71.5 & 66.0 & 64.7 & 65.9 & 62.5 & 45.2 & 45.4 & 41.3 \\
Qwen     & 63.7 & 55.3 & 58.8 & 71.0 & 61.3 & 64.0 & 43.1 & 44.5 & 41.9 \\
DeepSeek & 67.2 & 73.4 & 70.1 & 63.7 & 72.1 & 65.9 & 68.1 & 67.1 & 65.9 \\
Mistral  & 39.3 & 53.9 & 43.8 & 32.5 & 39.9 & 32.7 & 13.3 & 13.8 & 11.9 \\
Fanar    & 26.1 & 13.3 & 16.2 & 15.0 & 9.5 & 10.9 & 2.0 & 2.3 & 1.8 \\

\bottomrule
\end{tabular}
\caption{Per-dimension precision, recall, and F\textsubscript{1} at the Error Type, Correction Type, and Explanation Type levels under the \textbf{Full} setting on \textbf{QALB-2014}. \textbf{Bold} indicates the best score in each column.}
\label{tab:per-dim-qalb2014}
\end{table}

\begin{table}[th!]
\centering
\begin{tabular}{l ccc ccc ccc}
\toprule
& \multicolumn{3}{c}{\textbf{Error Type}} & \multicolumn{3}{c}{\textbf{Correction Type}} & \multicolumn{3}{c}{\textbf{Explanation Type}} \\
\cmidrule(lr){2-4} \cmidrule(lr){5-7} \cmidrule(lr){8-10}
\textbf{Model} & P & R & F\textsubscript{1} & P & R & F\textsubscript{1} & P & R & F\textsubscript{1} \\
\midrule
\multicolumn{10}{c}{\textbf{Lexical}} \\
\midrule
GPT      & 72.5 & 86.2 & 78.7 & 68.7 & 84.2 & 75.5 & 62.0 & 78.5 & 68.5 \\
Gemini   & 75.7 & \textbf{88.9} & 81.7 & \textbf{75.6} & \textbf{85.0} & \textbf{79.8} & \textbf{72.1} & \textbf{79.2} & \textbf{75.3} \\
Opus     & 66.0 & 77.3 & 71.0 & 58.6 & 68.3 & 62.4 & 43.9 & 56.0 & 48.4 \\
Qwen     & 64.1 & 72.5 & 67.9 & 69.5 & 69.6 & 68.5 & 50.0 & 58.3 & 51.5 \\
DeepSeek & \textbf{75.8} & \textbf{88.9} & \textbf{81.8} & 71.3 & 80.6 & 75.4 & 69.3 & 76.4 & 72.3 \\
Mistral  & 57.0 & 62.3 & 59.3 & 61.7 & 55.1 & 51.3 & 42.3 & 43.5 & 37.7 \\
Fanar    & 35.2 & 22.8 & 27.0 & 29.6 & 17.7 & 21.0 & 19.0 & 5.9 & 8.6 \\
\midrule
\multicolumn{10}{c}{\textbf{Orthography}} \\
\midrule
GPT      & 88.2 & 83.1 & 85.1 & 78.5 & \textbf{77.8} & 75.1 & 79.3 & 77.8 & 75.9 \\
Gemini   & \textbf{92.4} & \textbf{87.0} & \textbf{89.2} & \textbf{83.6} & 75.1 & \textbf{77.9} & \textbf{84.7} & \textbf{79.6} & \textbf{80.7} \\
Opus     & 77.5 & 68.1 & 69.7 & 45.7 & 40.5 & 42.2 & 41.3 & 35.6 & 36.3 \\
Qwen     & 84.2 & 67.7 & 66.7 & 41.8 & 32.9 & 34.4 & 34.7 & 29.3 & 29.6 \\
DeepSeek & 87.5 & 80.1 & 81.9 & 64.8 & 58.2 & 59.7 & 65.1 & 60.4 & 61.0 \\
Mistral  & 41.5 & 55.9 & 47.3 & 16.8 & 15.1 & 14.2 & 13.0 & 11.2 & 10.0 \\
Fanar    & 35.6 & 36.6 & 32.1 & 8.1 & 6.8 & 6.5 & 4.6 & 5.3 & 3.9 \\
\midrule
\multicolumn{10}{c}{\textbf{Morphology}} \\
\midrule
GPT      & 69.3 & 77.1 & 71.9 & 71.0 & 72.1 & 71.1 & 73.9 & 75.2 & 73.9 \\
Gemini   & \textbf{77.8} & \textbf{80.6} & \textbf{78.9} & \textbf{77.1} & \textbf{77.3} & \textbf{76.3} & \textbf{80.0} & \textbf{79.3} & \textbf{78.9} \\
Opus     & 64.9 & 56.8 & 58.9 & 56.2 & 49.9 & 51.9 & 44.6 & 41.5 & 42.5 \\
Qwen     & 60.6 & 57.9 & 58.0 & 44.6 & 43.8 & 43.3 & 44.5 & 42.5 & 42.4 \\
DeepSeek & 62.7 & 62.8 & 62.5 & 69.5 & 67.4 & 67.4 & 68.0 & 65.9 & 65.5 \\
Mistral  & 35.0 & 43.1 & 37.1 & 17.1 & 15.1 & 13.3 & 14.8 & 13.4 & 11.3 \\
Fanar    & 31.2 & 22.0 & 20.0 & 6.9 & 3.1 & 2.9 & 5.9 & 2.7 & 2.5 \\
\midrule
\multicolumn{10}{c}{\textbf{Syntax}} \\
\midrule
GPT      & 69.0 & 67.7 & 67.7 & 76.4 & 74.5 & 74.6 & 75.0 & 80.2 & 76.2 \\
Gemini   & \textbf{89.9} & \textbf{85.0} & \textbf{85.1} & \textbf{89.8} & \textbf{84.5} & \textbf{83.3} & \textbf{81.5} & \textbf{81.7} & \textbf{78.7} \\
Opus     & 49.9 & 71.8 & 51.4 & 45.6 & 48.8 & 43.9 & 31.2 & 31.3 & 28.2 \\
Qwen     & 78.2 & 68.1 & 71.7 & 69.6 & 53.9 & 57.3 & 43.1 & 40.9 & 39.5 \\
DeepSeek & 56.1 & 65.4 & 56.1 & 71.1 & 68.0 & 63.6 & 55.4 & 59.8 & 55.1 \\
Mistral  & 22.3 & 46.5 & 27.2 & 16.6 & 19.6 & 16.0 & 5.4 & 11.0 & 6.5 \\
Fanar    & 34.5 & 11.0 & 14.6 & 11.5 & 6.5 & 7.7 & 0.4 & 1.3 & 0.7 \\

\bottomrule
\end{tabular}
\caption{Per-dimension precision, recall, and F\textsubscript{1} at the Error Type, Correction Type, and Explanation Type levels under the \textbf{Full} setting on \textbf{QALB-2015}. \textbf{Bold} indicates the best score in each column.}
\label{tab:per-dim-qalb2015}
\end{table}

\begin{table}[th!]
\centering
\begin{tabular}{l ccc ccc ccc}
\toprule
& \multicolumn{3}{c}{\textbf{Error Type}} & \multicolumn{3}{c}{\textbf{Correction Type}} & \multicolumn{3}{c}{\textbf{Explanation Type}} \\
\cmidrule(lr){2-4} \cmidrule(lr){5-7} \cmidrule(lr){8-10}
\textbf{Model} & P & R & F\textsubscript{1} & P & R & F\textsubscript{1} & P & R & F\textsubscript{1} \\
\midrule
\multicolumn{10}{c}{\textbf{Lexical}} \\
\midrule
GPT      & \textbf{85.0} & 91.7 & 86.6 & \textbf{75.0} & 70.0 & 70.5 & 50.0 & 57.1 & 52.4 \\
Gemini   & 76.7 & 91.7 & 82.8 & 65.0 & 70.0 & 67.1 & 50.0 & 57.1 & 52.4 \\
Opus     & 70.7 & 91.7 & 79.6 & 57.8 & 52.8 & 53.7 & 41.2 & 43.8 & 42.4 \\
Qwen     & \textbf{85.0} & 83.3 & 81.2 & \textbf{75.0} & 65.0 & 67.6 & 50.0 & 53.6 & 50.3 \\
DeepSeek & 83.3 & \textbf{92.9} & \textbf{87.5} & 73.3 & \textbf{73.3} & \textbf{73.3} & \textbf{66.7} & \textbf{71.4} & \textbf{68.6} \\
Mistral  & 52.7 & 60.7 & 55.1 & 46.7 & 50.0 & 45.2 & 40.0 & 42.9 & 38.3 \\
Fanar    & 30.0 & 25.0 & 27.3 & 33.3 & 25.0 & 26.7 & 28.6 & 21.4 & 23.8 \\
\midrule
\multicolumn{10}{c}{\textbf{Orthography}} \\
\midrule
GPT      & 85.4 & \textbf{97.1} & 89.0 & \textbf{83.2} & \textbf{83.9} & \textbf{82.1} & \textbf{81.4} & \textbf{81.0} & \textbf{79.5} \\
Gemini   & \textbf{87.5} & 96.2 & \textbf{89.6} & 73.7 & 74.4 & 73.9 & 71.1 & 74.1 & 71.4 \\
Opus     & 83.9 & 95.2 & 86.8 & 44.8 & 44.4 & 44.5 & 42.9 & 43.2 & 43.0 \\
Qwen     & 58.3 & 67.3 & 61.7 & 54.0 & 54.8 & 53.2 & 43.5 & 47.4 & 43.3 \\
DeepSeek & 52.7 & 73.8 & 58.6 & 50.4 & 54.0 & 51.8 & 40.7 & 46.0 & 42.6 \\
Mistral  & 47.6 & 60.1 & 50.4 & 29.5 & 36.7 & 31.8 & 23.3 & 27.8 & 22.8 \\
Fanar    & 25.4 & 28.5 & 22.6 & 8.0 & 10.6 & 8.2 & 4.5 & 11.1 & 5.6 \\
\midrule
\multicolumn{10}{c}{\textbf{Morphology}} \\
\midrule
GPT      & \textbf{87.5} & \textbf{87.5} & \textbf{87.5} & \textbf{92.9} & \textbf{92.9} & \textbf{92.9} & \textbf{93.8} & \textbf{93.8} & \textbf{93.8} \\
Gemini   & 77.8 & 70.4 & 73.3 & 75.0 & 71.9 & 72.9 & 77.8 & 75.0 & 75.9 \\
Opus     & 65.8 & 60.0 & 61.0 & 48.3 & 42.5 & 44.0 & 48.5 & 45.5 & 46.1 \\
Qwen     & 54.2 & 57.3 & 55.5 & 43.5 & 43.1 & 42.2 & 43.2 & 42.0 & 41.8 \\
DeepSeek & 75.0 & 75.9 & 75.2 & 76.7 & 80.0 & 77.8 & 79.4 & 82.4 & 80.4 \\
Mistral  & 34.7 & 38.3 & 35.4 & 5.5 & 8.7 & 6.6 & 4.4 & 7.4 & 5.5 \\
Fanar    & 11.3 & 15.0 & 11.7 & 0.6 & 2.0 & 0.9 & 0.5 & 1.7 & 0.8 \\
\midrule
\multicolumn{10}{c}{\textbf{Syntax}} \\
\midrule
GPT      & \textbf{83.3} & \textbf{83.3} & \textbf{83.3} & \textbf{80.0} & \textbf{80.0} & \textbf{80.0} & \textbf{71.9} & \textbf{71.9} & \textbf{70.8} \\
Gemini   & \textbf{83.3} & \textbf{83.3} & \textbf{83.3} & \textbf{80.0} & \textbf{80.0} & \textbf{80.0} & 64.6 & 68.8 & 65.6 \\
Opus     & 66.7 & 78.5 & 69.6 & 47.2 & 50.0 & 48.3 & 17.7 & 19.4 & 17.5 \\
Qwen     & 44.3 & 40.3 & 42.1 & 38.3 & 37.1 & 37.7 & 32.4 & 29.4 & 30.4 \\
DeepSeek & \textbf{83.3} & 81.2 & 82.2 & 70.0 & 70.0 & 70.0 & \textbf{71.9} & 68.8 & 68.8 \\
Mistral  & 29.9 & 29.9 & 29.9 & 11.3 & 12.1 & 11.6 & 11.1 & 9.3 & 9.9 \\
Fanar    & 12.5 & 6.2 & 8.3 & 7.5 & 4.3 & 5.5 & 0.0 & 0.0 & 0.0 \\

\bottomrule
\end{tabular}
\caption{Per-dimension precision, recall, and F\textsubscript{1} at the Error Type, Correction Type, and Explanation Type levels under the \textbf{Full} setting on \textbf{ZAEBUC}. \textbf{Bold} indicates the best score in each column.}
\label{tab:per-dim-zaebuc}
\end{table}

\newpage
\clearpage
\section{Taxonomy Examples}
\label{app:taxonomy-examples}

\begin{figure*}[h]
    \centering
    \includegraphics[width=\textwidth]{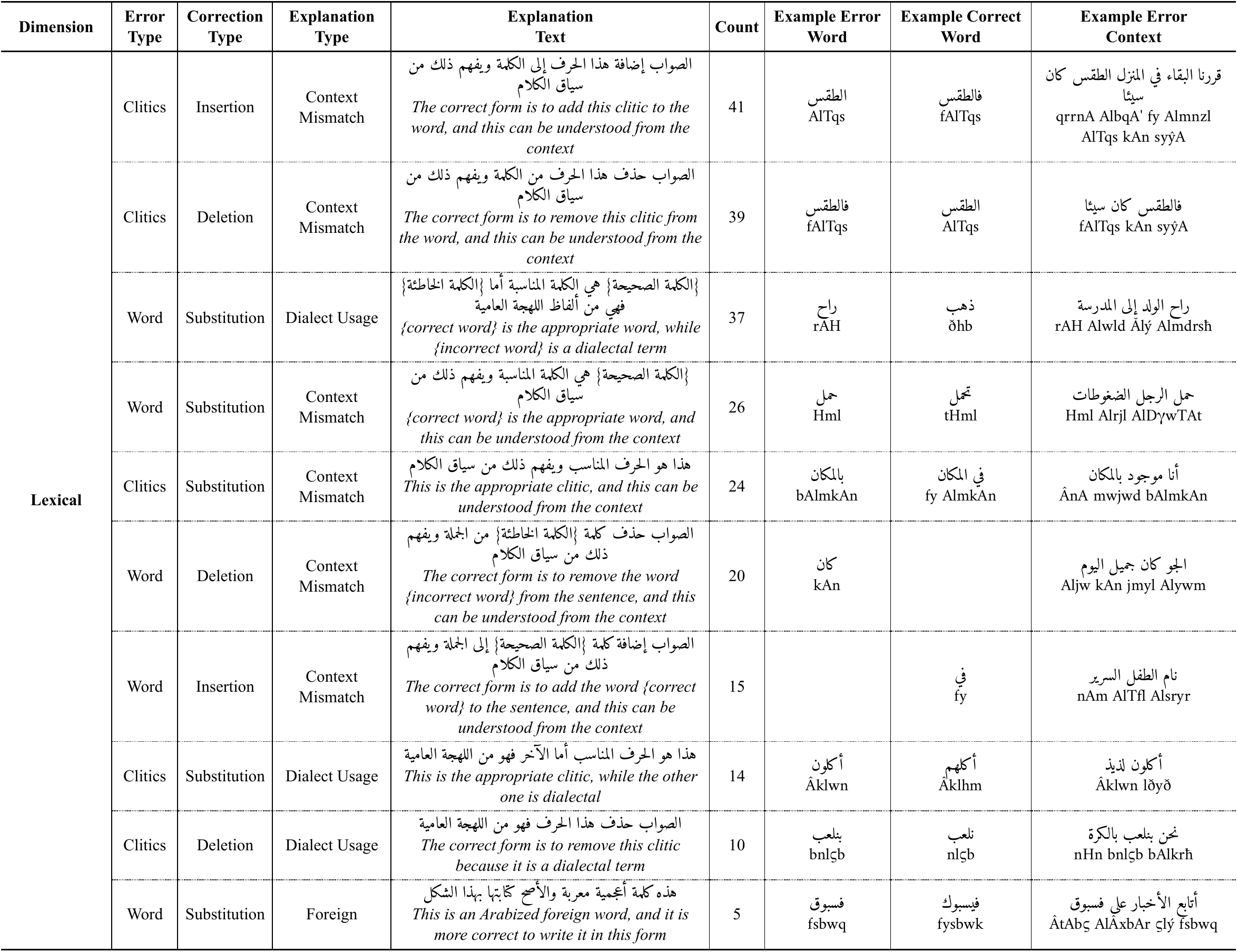}
    \caption{Top 10 \textbf{lexical} explanations in our annotated data, with their counts and representative examples.}
    \label{fig:lexical_explanations}
\end{figure*}

\clearpage
\begin{figure*}[t!]
    \centering
    \includegraphics[width=\textwidth]{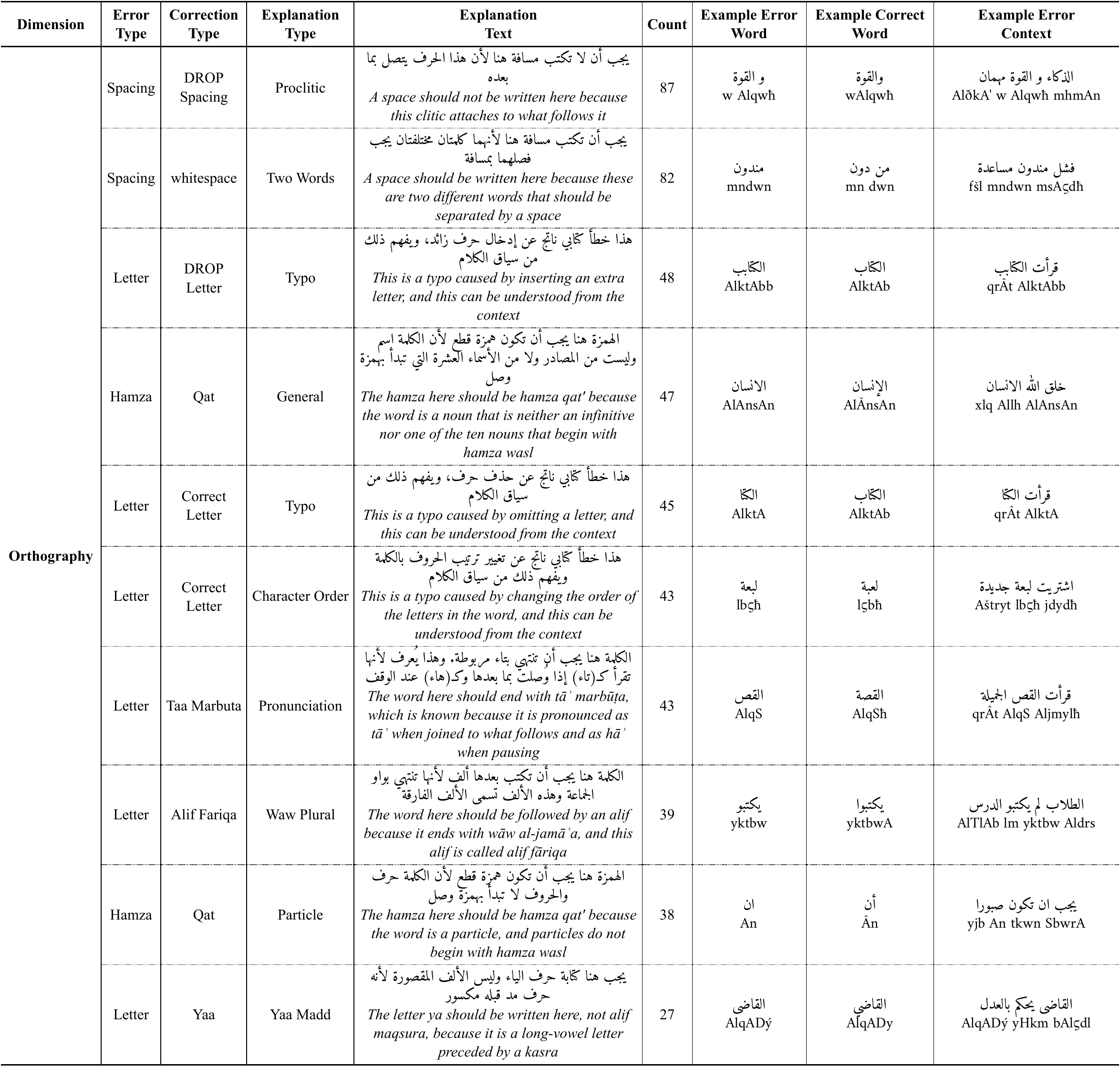}
    \caption{Top 10 \textbf{orthographic} explanations in our annotated data, with their counts and representative examples.}
    \label{fig:orthographic_explanations}
\end{figure*}

\clearpage

\begin{figure*}[t!]
    \centering
    \includegraphics[width=\textwidth]{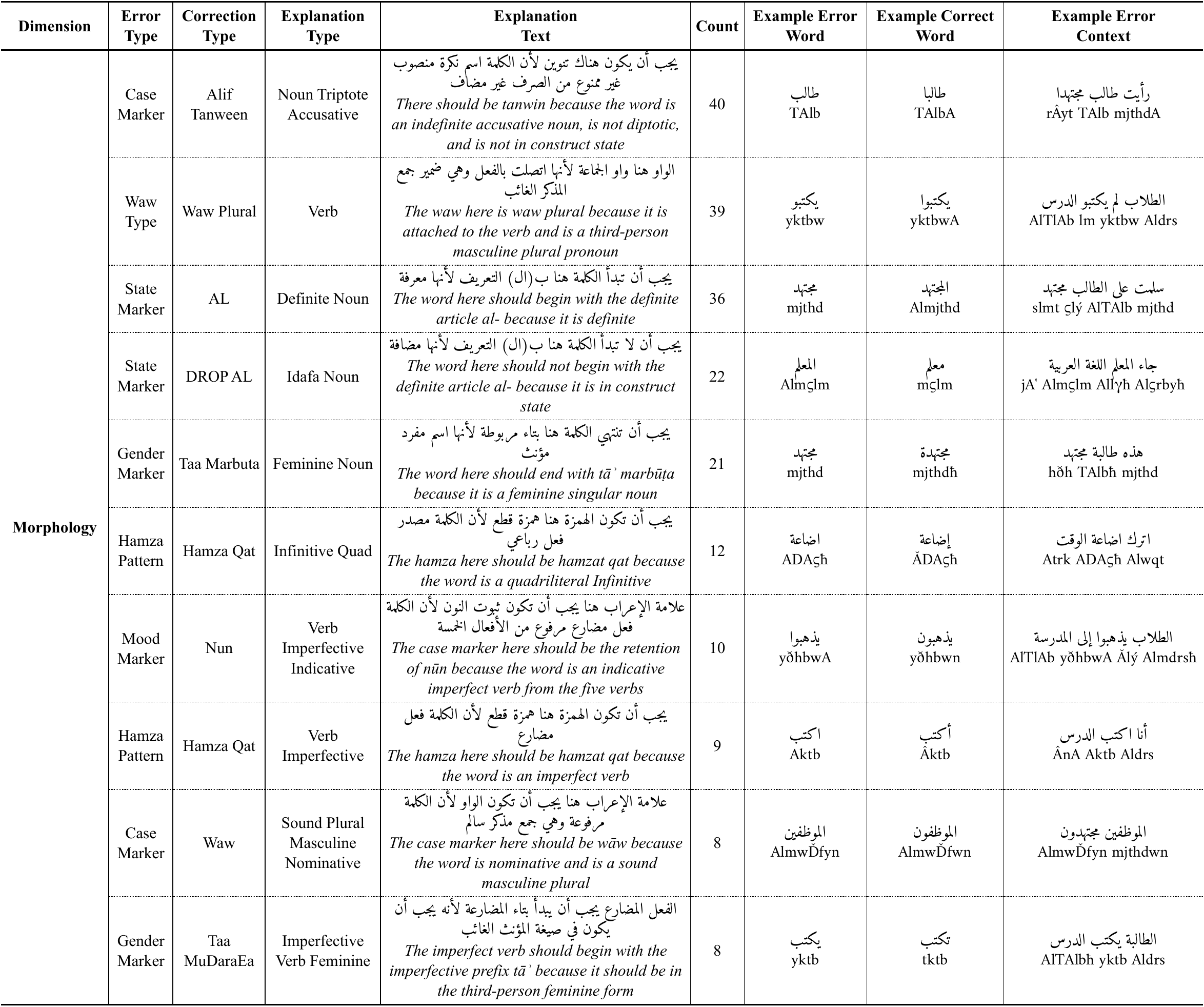}
    \caption{Top 10 \textbf{morphological} explanations in our annotated data, with their counts and representative examples.}
    \label{fig:morphological_explanations}
\end{figure*}

\begin{figure*}[t!]
    \centering
    \includegraphics[width=\textwidth]{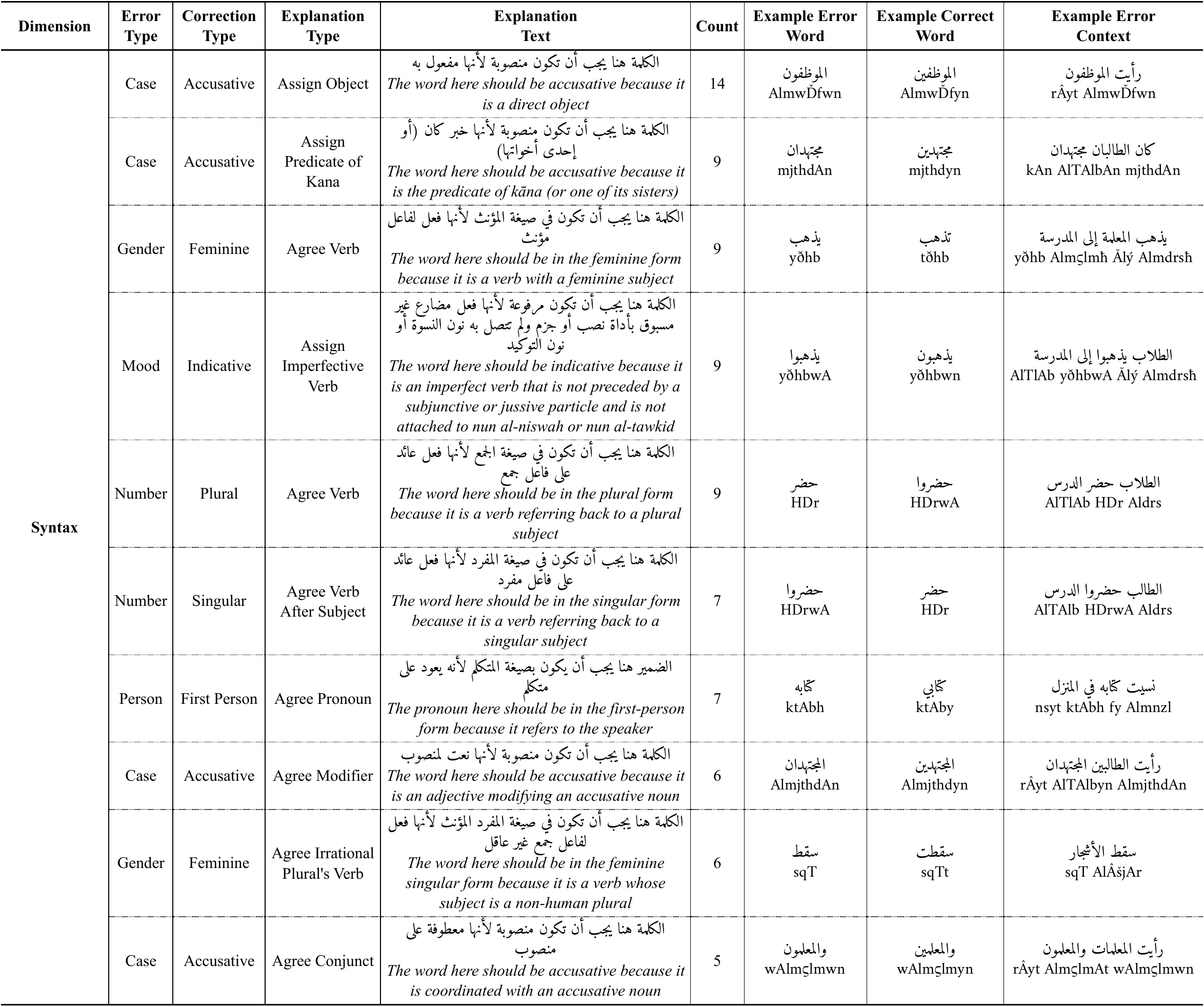}
    \caption{Top 10 \textbf{syntactic} explanations in our annotated data, with their counts and representative examples.}
    \label{fig:syntactic_explanations}
\end{figure*}

\clearpage


\end{document}